\documentclass[runningheads]{llncs}
\usepackage{times}
\usepackage{epsfig}
\usepackage{graphicx}
\usepackage{comment}
\usepackage{amsmath,amssymb} % define this before the line numbering.
\usepackage{color}
\usepackage{amsmath}
\usepackage{amssymb}
\usepackage{xcolor}
\usepackage{subcaption}
\usepackage{dblfloatfix}
\usepackage{setspace}
\usepackage{bbm}
\usepackage{gensymb}
\usepackage{booktabs}
\usepackage{cite}

\usepackage[ruled,lined,linesnumbered]{algorithm2e}
\usepackage[width=122mm,left=12mm,paperwidth=146mm,height=193mm,top=12mm,paperheight=217mm]{geometry}

\DeclareMathOperator*{\argmin}{arg\,min}

\renewcommand\vec{\mathbf}

\newcommand{\lblsec}[1]{\label{sec:#1}}
\newcommand{\lblfig}[1]{\label{fig:#1}} 
\newcommand{\lbltab}[1]{\label{tbl:#1}}

\newcommand{\refsec}[1]{Section~\ref{sec:#1}}
\newcommand{\reffig}[1]{Figure~\ref{fig:#1}} 
\newcommand{\reftab}[1]{Table~\ref{tbl:#1}}

\renewcommand{\paragraph}[1]{\noindent\textbf{#1}}

\begin{document}
% \renewcommand\thelinenumber{\color[rgb]{0.2,0.5,0.8}\normalfont\sffamily\scriptsize\arabic{linenumber}\color[rgb]{0,0,0}}
% \renewcommand\makeLineNumber {\hss\thelinenumber\ \hspace{6mm} \rlap{\hskip\textwidth\ \hspace{6.5mm}\thelinenumber}}
% \linenumbers
\pagestyle{headings}
\mainmatter
\def\ECCVSubNumber{****}  % Insert your submission number here

\title{Tracking Objects as Points} % Replace with your title

% INITIAL SUBMISSION 
\begin{comment}
\titlerunning{ECCV-20 submission ID \ECCVSubNumber} 
\authorrunning{ECCV-20 submission ID \ECCVSubNumber} 
\author{Anonymous ECCV submission}
\institute{Paper ID \ECCVSubNumber}
\end{comment}
%******************

% CAMERA READY SUBMISSION
% \begin{comment}
\titlerunning{Tracking Objects as Points}
% If the paper title is too long for the running head, you can set
% an abbreviated paper title here
%
\author{Xingyi Zhou\inst{1} \and
Vladlen Koltun\inst{2} \and
Philipp Kr\"ahenb\"uhl\inst{1}}
\authorrunning{Zhou et al.}
% First names are abbreviated in the running head.
% If there are more than two authors, 'et al.' is used.
%
\institute{$^{1}$UT Austin, $^{2}$Intel Labs}
% \end{comment}
%******************
\maketitle

\begin{abstract}
Tracking has traditionally been the art of following interest points through space and time. This changed with the rise of powerful deep networks. Nowadays, tracking is dominated by pipelines that perform object detection followed by temporal association, also known as tracking-by-detection. We present a simultaneous detection and tracking algorithm that is simpler, faster, and more accurate than the state of the art. Our tracker, CenterTrack, applies a detection model to a pair of images and detections from the prior frame. Given this minimal input, CenterTrack localizes objects and predicts their associations with the previous frame. That's it. CenterTrack is simple, online (no peeking into the future), and real-time. It achieves $67.8\%$ MOTA on the MOT17 challenge at 22 FPS and $89.4\%$ MOTA on the KITTI tracking benchmark at 15 FPS, setting a new state of the art on both datasets. CenterTrack is easily extended to monocular 3D tracking by regressing additional 3D attributes. Using monocular video input, it achieves $28.3\%$ AMOTA@0.2 on the newly released nuScenes 3D tracking benchmark, substantially outperforming the monocular baseline on this benchmark while running at 28 FPS.
\end{abstract}

\section{Introduction}

In early computer vision, tracking was commonly phrased as following interest points through space and time~\cite{tomasi1991detection,shi1994good}.
Early trackers were simple, fast, and reasonably robust.
However, they were liable to fail in the absence of strong low-level cues such as corners and intensity peaks.
With the advent of high-performing object detection models~\cite{felzenszwalb2009object,ren2015faster}, a powerful alternative emerged: tracking-by-detection (or more precisely, tracking-after-detection)~\cite{Bewley2016_sort,tang2017multiple,xu2019spatial}.
These models rely on a given accurate recognition to identify objects and then link them up through time in a separate stage.
Tracking-by-detection leverages the power of deep-learning-based object detectors and is currently the dominant tracking paradigm. Yet the best-performing object trackers are not without drawbacks. Many rely on slow and complex association strategies to link detected boxes through time~\cite{xiang2015learning,Hu3DT19,tang2017multiple,xu2019spatial}. 
Recent work on \emph{simultaneous detection and tracking}~\cite{bergmann2019tracking,feichtenhofer2017detect} has made progress in alleviating some of this complexity.
Here, we show how combining ideas from point-based tracking and simultaneous detection and tracking further simplifies tracking.

\begin{figure}[t]
\captionsetup{type=figure}
\centering
  \begin{subfigure}[t]{0.25\linewidth}
   \includegraphics[width=0.99\linewidth,page=1]{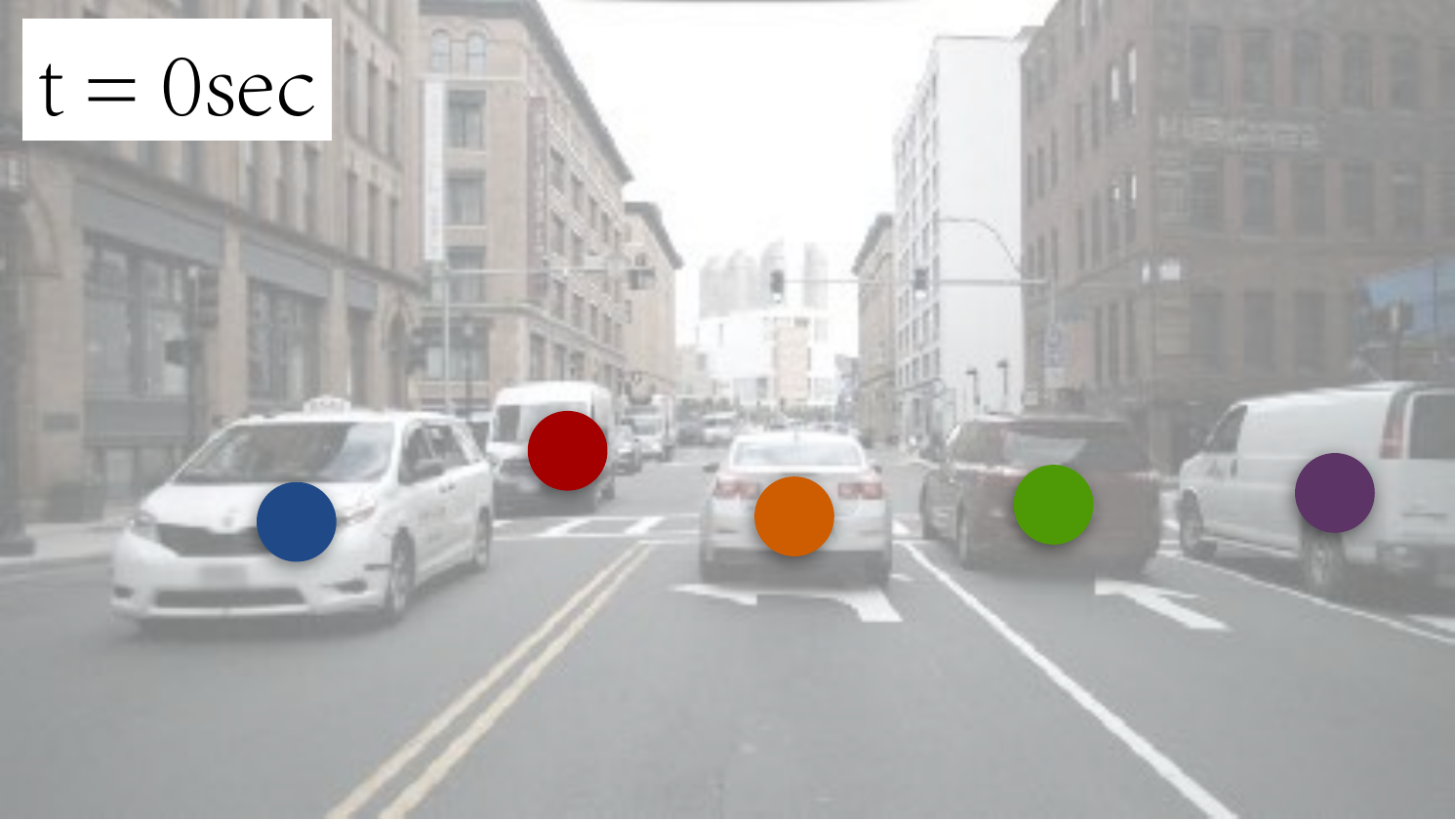}
  \end{subfigure}%
  \begin{subfigure}[t]{0.25\linewidth}
   \includegraphics[width=0.99\linewidth,page=2]{figs/teaser-v5.pdf}
  \end{subfigure}%
  \begin{subfigure}[t]{0.25\linewidth}
   \includegraphics[width=0.99\linewidth,page=3]{figs/teaser-v5.pdf}
  \end{subfigure}%
  \begin{subfigure}[t]{0.25\linewidth}
   \includegraphics[width=0.99\linewidth,page=4]{figs/teaser-v5.pdf}
  \end{subfigure}%
   \caption{We track objects by tracking their centers. We learn a 2D offset between two adjacent frames and associate them based on center distance.}
   \lblfig{teaser}
\vspace{-7mm}
\end{figure}

We present a point-based framework for joint detection and tracking, referred to as CenterTrack.
Each object is represented by a single point at the center of its bounding box.
This center point is then tracked through time (\reffig{teaser}).
Specifically, we adopt the recent CenterNet detector to localize object centers~\cite{zhou2019objects}.
We condition the detector on two consecutive frames, as well as a heatmap of prior tracklets, represented as points.
We train the detector to also output an offset vector from the current object center to its center in the previous frame.
We learn this offset as an attribute of the center point at little additional computational cost.
A greedy matching, based solely on the distance between this predicted offset and the detected center point in the previous frame, suffices for object association.
The tracker is end-to-end trainable and differentiable.

Tracking objects as points simplifies two key components of the tracking pipeline.
First, it simplifies tracking-conditioned detection.
If each object in past frames is represented by a single point, a constellation of objects can be represented by a heatmap of points~\cite{carreira2016human}.
Our tracking-conditioned detector directly ingests this heatmap and reasons about all objects jointly when associating them across frames.
Second, point-based tracking simplifies object association across time.
A simple displacement prediction, akin to sparse optical flow, allows objects in different frames to be linked.
This displacement prediction is conditioned on prior detections.
It learns to jointly detect objects in the current frame and associate them to prior detections.

While the overall idea is simple, subtle details matter in making this work.
Tracked objects in consecutive frames are highly correlated.
With the previous-frame heatmap given as input, CenterTrack could easily learn to repeat the predictions from the preceding frame, and thus refuse to track without incurring a large training error.
We prevent this through an aggressive data-augmentation scheme during training.
In fact, our data augmentation is aggressive enough for the model to learn to track objects from \emph{static images}. That is, CenterTrack can be successfully trained on static image datasets (with ``hallucinated'' motion), with no real video input.

CenterTrack is purely local. It only associates objects in adjacent frames, without reinitializing lost long-range tracks.
It trades the ability to reconnect long-range tracks for simplicity, speed, and high accuracy in the local regime.
Our experiments indicate that this trade-off is well worth it.
CenterTrack outperforms complex tracking-by-detection strategies on the MOT~\cite{MOT16} and KITTI~\cite{Geiger2012CVPR} tracking benchmarks.
We further apply the approach to monocular 3D object tracking on the nuScenes dataset~\cite{nuscenes2019}.
Our monocular tracker achieves $28.3\%$ AMOTA@0.2, outperforming the monocular baseline by a factor of 3, while running at 22 FPS.
It can be trained on labelled video sequences, if available, or on static images with data augmentation.
Code is available at \textcolor{magenta}{\url{https://github.com/xingyizhou/CenterTrack}}.

\section{Related work}

\paragraph{Tracking-by-detection.}
Most modern trackers ~\cite{Bewley2016_sort,Wojke2017simple,tang2017multiple,xu2019spatial,sharma2018beyond,zhu2018online,leal2016learning,fang2018recurrent,schulter2017deep} follow the tracking-by-detection paradigm.
An off-the-shelf object detector~\cite{ren2015faster,felzenszwalb2009object,yang2016exploit,ren2017accurate} first finds all objects in each individual frame.
Tracking is then a problem of bounding box association.
SORT~\cite{Bewley2016_sort} tracks bounding boxes using a Kalman filter and associates each bounding box with its highest overlapping detection in the current frame using bipartite matching.
DeepSORT~\cite{Wojke2017simple} augments the overlap-based association cost in SORT with appearance features from a deep network.
More recent approaches focus on increasing the robustness of object association.
Tang et al.~\cite{tang2017multiple} leverage person-reidentification features and human pose features.
Xu et al.~\cite{xu2019spatial} take advantage of the spatial locations over time.
BeyondPixel~\cite{sharma2018beyond} uses additional 3D shape information to track vehicles.

These methods have two drawbacks. First, the data association discards image appearance features~\cite{Bewley2016_sort} or requires a computationally expensive feature extractor~\cite{tang2017multiple, sharma2018beyond,xu2019spatial,feng2019multi}. Second, detection is separated from tracking.
In our approach, association is almost free.
Association is \emph{learned} jointly with detection.
Also, our detector takes the previous tracking results as an input, and can learn to recover missing or occluded objects from this additional cue.

\paragraph{Joint detection and tracking.}
A recent trend in multi-object tracking is to convert existing detectors into trackers and combine both tasks in the same framework.
Feichtenhofer et al.~\cite{feichtenhofer2017detect} use a siamese network with the current and past frame as input and predict inter-frame offsets between bounding boxes.
Integrated detection~\cite{zhang2018integrated} uses tracked bounding boxes as additional region proposals to enhance detection, followed by bipartite-matching-based bounding-box association.
Tracktor~\cite{bergmann2019tracking} removes the box association
by directly propagating identities of region proposals using bounding box regression.
In video object detection, Kang et al.~\cite{kang2017object,kang2017t} feed stacked consecutive frames into the network and do detection for a whole video segment. 
And Zhu et al.~\cite{zhu2017flow} use flow to warp intermediate features from previous frames to accelerate inference.

Our method belongs to this category. The difference is that all of these works adopt the FasterRCNN framework~\cite{ren2015faster}, where the tracked boxes are used as region proposals.
This assumes that bounding boxes have a large overlap between frames, which is not true in low-framerate regimes.
As a consequence, Tracktor~\cite{bergmann2019tracking} requires a motion model~\cite{evangelidis2008parametric,choi2010multiple} for low-framerate sequences. 
Our approach instead provides the tracked predictions as an additional point-based heatmap input to the network.
The network is then able to reason about and match objects anywhere in its receptive field even if the boxes 
have no overlap at all.

\paragraph{Motion prediction.}
Motion prediction is another important component in a tracking system.
Early approaches~\cite{Bewley2016_sort,Wojke2017simple} used Kalman filters to model object velocities.
Held et al.~\cite{held2016learning} use a regression network to predict four scalars for bounding box offset between frames for single-object tracking.
Xiao et al.~\cite{xiao2018simple} utilize an optical flow estimation network to update joint locations in human pose tracking.
Voigtlaender et al.~\cite{voigtlaender2019mots} learn a high-dimensional embedding vector for object identities for simultaneous object tracking and segmentation.
Our center offset is analogous to sparse optical flow, but is learned together with the detection network and does not require dense supervision.

\paragraph{Heatmap-conditioned keypoint estimation.}
Feeding the model predictions as an additional input to a model works across a wide range of vision tasks~\cite{tu2008auto}, especially for keypoint estimation~\cite{carreira2016human,fieraru2018learning,Moon_2019_CVPR_PoseFix}. 
Auto-context~\cite{tu2008auto} feeds the mask prediction back into the network.
Iterative-Error-Feedback (IEF)~\cite{carreira2016human} takes another step by rendering predicted keypoint coordinates into heatmaps.
PoseFix~\cite{Moon_2019_CVPR_PoseFix} generates heatmaps that simulate test errors for human pose refinement.

Our tracking-conditioned detection framework is inspired by these works.
A rendered heatmap of prior keypoints~\cite{tu2008auto,carreira2016human,fieraru2018learning,Moon_2019_CVPR_PoseFix} is especially appealing in tracking for two reasons. First, the information in the previous frame is freely available and does not slow down the detector.
Second, conditional tracking can reason about occluded objects that may no longer be visible in the current frame. The tracker can simply learn to keep those detections from the prior frame around.

\paragraph{3D object detection and tracking.}
3D trackers replace the object detection component in standard tracking systems with 3D detection from monocular images~\cite{ren2017accurate} or 3D point clouds~\cite{zhu2019class,shi2019pointrcnn}.
Tracking then uses an off-the-shelf identity association model.
For example, 3DT~\cite{Hu3DT19} detects 2D bounding boxes, estimates 3D motion, and uses depth and order cues for matching.
AB3D~\cite{Weng2019_3dmot} achieves state-of-the-art performance by combining a Kalman filter with accurate 3D detections~\cite{shi2019pointrcnn}.

\section{Preliminaries}

Our method, CenterTrack, builds on the CenterNet detector~\cite{zhou2019objects}.
CenterNet takes a single image $I \in \mathbb{R}^{W \times H \times 3}$ as input and produces a set of detections $\{(\vec p_i, \vec s_i)\}_{i=0}^{N-1}$ for each class $c \in \{0,\ldots,C-1\}$.
CenterNet identifies each object through its center point $\vec p \in \mathbb{R}^2$ and then regresses to a height and width $\vec s \in \mathbb{R}^2$ of the object's bounding box.
Specifically, it produces a low-resolution heatmap $\hat{Y} \in [0, 1]^{\frac{W}{R} \times \frac{H}{R} \times C}$ and a size map $\hat{S} \in \mathbb{R}^{\frac{W}{R} \times \frac{H}{R} \times 2}$ with a downsampling factor $R=4$. 
Each local maximum $\hat{\vec p} \in \mathbb{R}^2$ (also called peak, whose response is the strongest in a $3\times3$ neighborhood) in the heatmap $\hat{Y}$ corresponds to a center of a detected object with confidence $\hat w = \hat{Y}_{\hat{\vec p}}$ and object size $\hat{\vec s} = \hat{S}_{\hat{\vec p}}$.

Given an image with a set of annotated objects $\{\vec p_0, \vec p_1, \ldots\}$, CenterNet uses a training objective based on the focal loss~\cite{law2018cornernet,lin2017focal}:
\begin{equation}
    L_{k} = \frac{1}{N} \sum_{xyc}
    \begin{cases}
        (1 - \hat{Y}_{xyc})^{\alpha} 
        \log(\hat{Y}_{xyc}) & \!\text{if}\ Y_{xyc}=1\vspace{2mm}\\
        \begin{array}{c}
        (1-Y_{xyc})^{\beta} 
        (\hat{Y}_{xyc})^{\alpha}\log(1-\hat{Y}_{xyc})
        \end{array}
        & \!\text{otherwise}
    \end{cases},
\end{equation}
where $Y \in [0, 1]^{\frac{W}{R} \times \frac{H}{R} \times C}$ is a ground-truth heatmap corresponding to the annotated objects.
$N$ is the number of objects, and $\alpha=2$ and $\beta=4$ are hyperparameters of the focal loss. 
For each center $\vec p$ of class $c$, we render a Gaussian-shaped peak into $Y_{:,:,c}$ using a rendering function $Y = \mathcal{R}(\{\vec p_0, \vec p_1, \ldots\})$~\cite{law2018cornernet}.
Formally, the rendering function at position $\vec q \in \mathbb{R}^2$ is defined as
$$
\mathcal{R}_{\vec q}(\{\vec p_0, \vec p_1, \ldots\}) = \max_{i} \exp\left(-\frac{(\vec p_i-\vec q)^2}{2\sigma^2_i}\right).
$$
The Gaussian kernel $\sigma_i$ is a function of the object size~\cite{law2018cornernet}.

The size prediction is only supervised at the center locations.
Let $\vec s_i$ be the bounding box size of the $i$-th object at location $\vec p_i$.
Size prediction is learned by regression
\begin{equation}
L_{size} = \frac{1}{N}\sum_{i=1}^{N}|\hat{S}_{\vec p_i} - \vec s_i|.
\end{equation}

CenterNet further regresses to a refined center local location using an analogous L1 loss $L_{loc}$.
The overall loss of CenterNet is a weighted sum of all three loss terms: focal loss, size, and local location regression.

\begin{figure}[t]
\captionsetup{type=figure}
  \begin{subfigure}[t]{0.14\linewidth}
   \includegraphics[width=0.95\linewidth,page=2]{figs/fig2-new.pdf}
   \caption*{\scriptsize Image $I^{(t)}$}
  \end{subfigure}%
  \begin{subfigure}[t]{0.14\linewidth}
   \centering
   \includegraphics[width=0.95\linewidth,page=1]{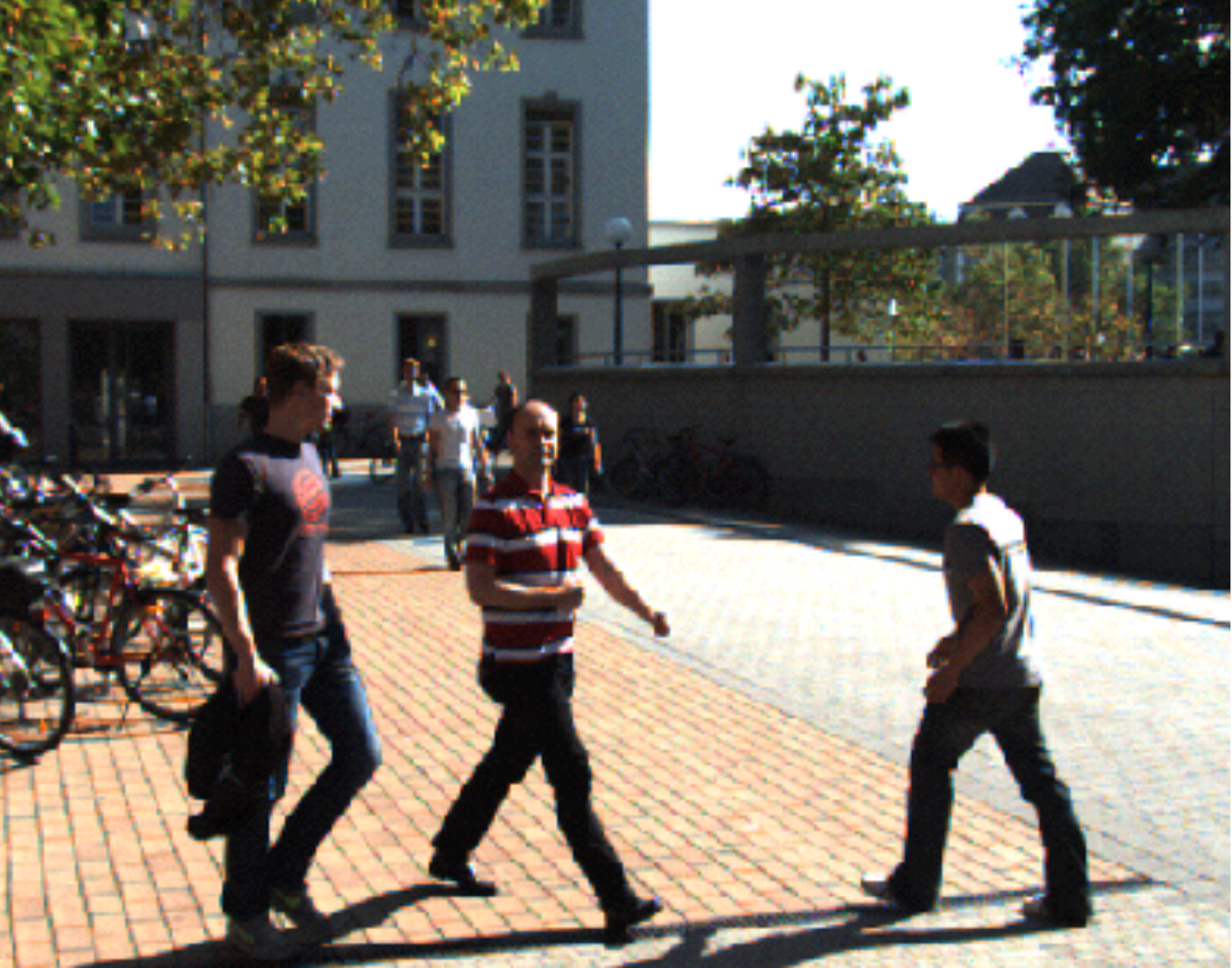}
   \caption*{\scriptsize Image $I^{(t-1)}$}
  \end{subfigure}%
  \begin{subfigure}[t]{0.14\linewidth}
   \centering
   \includegraphics[width=0.99\linewidth,page=3]{figs/fig2-new.pdf}
   \caption*{\scriptsize Tracks $T^{(t-1)}$}
  \end{subfigure}
  \begin{subfigure}[t]{0.14\linewidth}
   \centering
   \includegraphics[width=\linewidth,page=4]{figs/fig2-new.pdf}
   \caption*{}
  \end{subfigure}%
  \begin{subfigure}[t]{0.14\linewidth}
   \centering
   \includegraphics[width=0.95\linewidth,page=5]{figs/fig2-new.pdf}
   \caption*{\scriptsize Detections $\hat Y^{(t)}$}
  \end{subfigure}%
  \begin{subfigure}[t]{0.14\linewidth}
   \centering
   \includegraphics[width=0.95\linewidth,page=6]{figs/fig2-new.pdf}
   \caption*{\scriptsize Size $\hat S^{(t)}$}
  \end{subfigure}
  \begin{subfigure}[t]{0.14\linewidth}
   \centering
   \includegraphics[width=0.95\linewidth,page=7]{figs/fig2-new.pdf}
   \caption*{\scriptsize Offset $\hat O^{(t)}$}
  \end{subfigure}%
   \caption{Illustration of our framework. The network takes the current frame, the previous frame, and a heatmap rendered from tracked object centers as inputs, and produces a center detection heatmap for the current frame, the bounding box size map, and an offset map. At test time, object sizes and offsets are extracted from peaks in the heatmap.}
\lblfig{framework}
\end{figure}

\section{Tracking objects as points}
We approach tracking from a local perspective. 
When an object leaves the frame or is occluded and reappears, it is assigned a new identity.
We thus treat tracking as the problem of propagating detection identities across \emph{consecutive} frames, without re-establishing associations across temporal gaps.

At time $t$, we are given an image of the current frame $I^{(t)} \in \mathbb{R}^{W \times H \times 3}$ and the previous frame $I^{(t - 1)} \in \mathbb{R}^{W \times H \times 3}$, as well as the tracked objects in the previous frame $T^{(t - 1)} = \{b_{0}^{(t - 1)}, b_{1}^{(t - 1)}, \ldots\}_i$.
Each object $b = (\vec p, \vec s, w, id)$ is described by its center location $\vec p \in \mathbb{R}^2$, size $\vec s \in \mathbb{R}^2$, detection confidence $w \in [0, 1]$, and unique identity $id \in \mathbb{I}$.
Our aim is to detect and track objects $T^{(t)} = \{b_{0}^{(t)}, b_{1}^{(t)}, \ldots\}$ in the current frame $t$, and assign objects that appear in both frames a consistent $id$.

There are two main challenges here.
The first is finding all objects in every frame~-- including occluded ones.
The second challenge is associating these objects through time.
We address both via a single deep network, trained end-to-end.
\refsec{cond_track} describes a tracking-conditioned detector that leverages tracked detections from the previous frame to improve detection in the current frame.
\refsec{offset} then presents a simple offset prediction scheme that is able to link detections through time.
Finally, Sections \ref{sec:train_video} and \ref{sec:train_image} show how to train this detector from video or static image data.

\subsection{Tracking-conditioned detection}
\lblsec{cond_track}

As an object detector, CenterNet already infers most of the required information for tracking: object locations $\hat{\vec p}$, their size $\hat{\vec s} = \hat S_{\hat{ \vec p}}$, and a confidence measure $\hat w = \hat Y_{\hat{ \vec p}}$.
However, it is unable to find objects that are not directly visible in the current frame, and the detected objects may not be temporally coherent.
One natural way to increase temporal coherence is to provide the detector with additional image inputs from past frames.
In CenterTrack, we provide the detection network with two frames as input: the current frame $I^{(t)}$ and the prior frame $I^{(t-1)}$.
This allows the network to estimate the change in the scene and potentially recover occluded objects at time $t$ from visual evidence at time $t-1$.

CenterTrack also takes prior detections $\{\vec p^{(t-1)}_0, \vec p^{(t-1)}_1, \ldots\}$ as additional input.
How should these detections be represented in a form that is easily provided to a network?
The point-based nature of our tracklets is helpful here.
Since each detected object is represented by a single point, we can conveniently render all detections in a class-agnostic single-channel heatmap $H^{(t-1)}=\mathcal{R}(\{\vec p^{(t-1)}_0, \vec p^{(t-1)}_1, \ldots\})$, using the same Gaussian render function as in the training of point-based detectors.
To reduce the propagation of false positive detections, we only render objects with a confidence score greater than a threshold $\tau$.
The architecture of CenterTrack is essentially identical to CenterNet, with four additional input channels. (See \reffig{framework}.)

Tracking-conditioned detection provides a temporally coherent set of detected objects.
However, it does not link these detections across time.
In the next section, we show how to add one additional output to point-based detection to track objects through space and time.

\subsection{Association through offsets}
\lblsec{offset}
To associate detections through time, CenterTrack predicts a 2D displacement as two additional output channels $\hat D^{(t)} \in \mathbb{R}^{\frac{W}{R} \times \frac{H}{R} \times 2}$.
For each detected object at location $\hat{\vec p}^{(t)}$, the displacement $\hat{\vec d}^{(t)} = \hat D^{(t)}_{\hat{\vec p}^{(t)}}$ captures the difference in location of the object in the current frame $\hat{\vec p}^{(t)}$ and the previous frame $\hat{\vec p}^{(t-1)}$: $\hat{\vec d}^{(t)} = \hat{\vec p}^{(t)} - \hat{\vec p}^{(t-1)}$.
We learn this displacement using the same regression objective as size or location refinement:
\begin{equation}
L_{off} = \frac{1}{N}\sum_{i=1}^{N}\left|\hat{D}_{\vec{p}_i^{(t)}} - (\vec{p}_i^{(t-1)} - \vec{p}_i^{(t)})\right|,
\end{equation}
where $\vec{p}_i^{(t-1)}$ and $\vec{p}_i^{(t)}$ are tracked ground-truth objects.
\reffig{framework} shows an example of this offset prediction.

With a sufficiently good offset prediction, a simple greedy matching algorithm can associate objects across time.
For each detection at position $\hat p$, we greedily associate it with the closest unmatched prior detection at position $\hat p - \hat D_{\hat p}$, in descending order of confidence $\hat w$.
If there is no unmatched prior detection within a radius $\kappa$, we spawn a new tracklet.
We define $\kappa$ as the geometric mean of the width and height of the predicted bounding box for each tracklet.
A precise description of this greedy matching algorithm is provided in supplementary material.
The simplicity of this greedy matching algorithm again highlights the advantages of tracking objects as points.
A simple displacement prediction is sufficient to link objects across time.
There is no need for a complicated distance metric or graph matching.

\subsection{Training on video data}
\lblsec{train_video}

CenterTrack is first and foremost an object detector, and trained as such.
The architectural changed from CenterNet to CenterTrack are minor: four additional input channels and two output channels.
This allows us to fine-tune CenterTrack directly from a pretrained CenterNet detector~\cite{zhou2019objects}.
We copy all weights related to the current detection pipeline. All weights corresponding to additional inputs or outputs are initialized randomly.
We follow the CenterNet training protocol and train all predictions as multi-task learning.
We use the same training objective with the addition of offset regression $L_{off}$.

The main challenge in training CenterTrack comes in producing a realistic tracklet heatmap $H^{(t-1)}$.
At inference time, this tracklet heatmap can contain an arbitrary number of missing tracklets, wrongly localized objects, or even false positives.
These errors are not present in ground-truth tracklets $\{\vec p^{(t-1)}_0, \vec p^{(t-1)}_1, \ldots\}$ provided during training.
We instead simulate this test-time error during training.
Specifically, we simulate three types of error.
First, we locally jitter each tracklet $\vec p^{(t-1)}$ from the prior frame by adding Gaussian noise to each center. That is, we render $p_i' = (x_i + r \times \lambda_{jt} \times w_i, y_i + r \times \lambda_{jt} \times h_i)$, where r is sampled from a Gaussian distribution. We use $\lambda_{jt}=0.05$ in all experiments. 
Second, we randomly add false positives near ground-truth object locations by rendering a spurious noisy peak $p_i'$ with probability $\lambda_{fp}$.
Third, we simulate false negatives by randomly removing detections with probability $\lambda_{fn}$.
$\lambda_{fp}$ and $\lambda_{fn}$ are set according to the statistics of our baseline model.
These three augmentations are sufficient to train a robust tracking-conditioned object detector.

In practice, $I^{(t-1)}$ does not need to be the immediately preceding frame from time $t-1$. 
It can be a different frame from the same video sequence.
In our experiments, we randomly sample frames near $t$ to avoid overfitting to the framerate. Specifically, we sample from all frames $k$ where $|k - t| < M_f$, where $M_f=3$ is a hyperparameter.

\vspace{-4mm}

\subsection{Training on static image data}
\lblsec{train_image}
\label{sec:trainimage}
Without labeled video data, CenterTrack does not have access to a prior frame $I^{(t-1)}$ or tracked detections $\{\vec p^{(t-1)}_0, \vec p^{(t-1)}_1, \ldots\}$.
However, we can simulate tracking on standard detection benchmarks, given only single images $I^{(t)}$ and detections $\{\vec p^{(t)}_0, \vec p^{(t)}_1, \ldots\}$.
The idea is simple: we simulate the previous frame by randomly scaling and translating the current frame. As our experiments will demonstrate, this is surprisingly effective.

\vspace{-4mm}

\subsection{End-to-end 3D object tracking}
To perform monocular 3D tracking, we adopt the monocular 3D detection form of CenterNet~\cite{zhou2019objects}.
Specifically, we train output heads to predict object depth, rotation (encoded as an $8$-dimensional vector~\cite{Hu3DT19}), and 3D extent. 
Since the projection of the center of the 3D bounding box may not align with the center of the object's 2D bounding box (due to perspective projection), we also predict a 2D-to-3D center offset.
Further details are provided in the supplement.

\section{Experiments}

We evaluate 2D multi-object tracking on the MOT17~\cite{MOT16} and KITTI~\cite{Geiger2012CVPR} tracking benchmarks. We also evaluate monocular 3D tracking on the nuScenes dataset~\cite{nuscenes2019}. Experiments on MOT16 can be found in the supplement.

\subsection{Datasets and evaluation metrics}
\lblsec{datasets}
\paragraph{MOT.}
MOT17 contains 7 training sequences and 7 test sequences~\cite{MOT16},
The videos were captured by stationary cameras mounted in high-density scenes with heavy occlusion.
Only pedestrians are annotated and evaluated.
The video framerate is 25-30 FPS.
The MOT dataset does not provide an official validation split.
For ablation experiments, we split each training sequence into two halves, and use the first half frames for training and the second for validation.
Our main results are reported on the test set.

\paragraph{KITTI.}
The KITTI tracking benchmark consists of 21 training sequences and 29 test sequences~\cite{Geiger2012CVPR}. 
They are collected by a camera mounted on a car moving through traffic. The dataset provides 2D bounding box annotations for cars, pedestrians, and cyclists, but only cars are evaluated. 
Videos are captured at 10 FPS and contain large inter-frame motions.
KITTI does not provide detections, and all entries use private detection.
We again split all training sequences into halves for training and validation.

\paragraph{nuScenes.} nuScenes is a newly released large-scale driving dataset with 7 object classes annotated for tracking~\cite{nuscenes2019}.
It contains 700 training sequences, 150 validation sequences, and 150 test sequences. 
Each sequence contains roughly 40 frames at 2 FPS with 6 slightly overlapping images in a panoramic $360\degree$ view, resulting in 168k training, 36k validation, and 36k test images.
The videos are sampled at 12 FPS, but frames are only annotated and evaluated at 2 FPS.
All baselines and CenterTrack only use keyframes for training and evaluation. Due to the low framerate, the inter-frame motion is significant.

\paragraph{Evaluation metrics.}
We use the official evaluation metrics in each dataset.
The common metric is multi-object tracking accuracy~\cite{stiefelhagen2006clear, leal2017tracking}:
$MOTA = 1 - \frac{\sum_t(FP_t + FN_t + IDSW_t)}{\sum_t GT_t}$, where $GT_t$, $FP_t$, $FN_t$, and $IDSW_t$ are the number of ground-truth bounding boxes, false positives, false negatives, and identity switches in frame $t$, respectively.
MOTA does not rank tracklets according to confidence and is sensitive to the task-dependent output threshold  $\theta$~\cite{Weng2019_3dmot}.
The thresholds we use are listed in \refsec{implementation}.
The interplay between output threshold and true positive criteria matters.
For 2D tracking~\cite{MOT16, Geiger2012CVPR}, $>0.5$ bounding box IoU is a the true positive.
For 3D tracking~\cite{nuscenes2019}, bounding box center distance $<2m$ on the ground plane is the criterion for a true positive.
When objects are successfully detected, but not tracked, they are identified as an identity switch (IDSW).
The IDF1 metric measures the minimal cost change from predicted ids to the correct ids. 
In our ablation studies, we report false positve rate (FP) $\frac{\sum_t FP_t}{\sum_t GT_t}$, false negative rate (FN) $\frac{\sum_t FN_t}{\sum_t GT_t}$, and identity switches (IDSW) $\frac{\sum_t IDSW_t}{\sum_t GT_t}$ separately.
In comparisons with other methods, we report the absolute numbers following the dataset convention~\cite{MOT16,Geiger2012CVPR}.
We also report the Most Tracked ratio (MT) for the ratio of most tracked ($>80\%$ time) objects and Most Lost ratio (ML) for most lost ($<20\%$ time) objects~\cite{stiefelhagen2006clear}.

nuScenes adopts a more robust metric, AMOTA, which is a weighted average of MOTA across different output thresholds. 
Specifically,
{
\scriptsize
\begin{align*}
& AMOTA = \frac{1}{n-1} \sum_{r \in \{\frac{1}{n-1}, \frac{2}{n-1}, \cdots, 1\}} MOTA_r \\
& MOTA_r = max(0, 1 - \alpha \frac{IDSW_r + FP_r + FN_r - (1 - r)\times P}{r \times P}) \\
\end{align*}
}%
where $r$ is a fixed recall threshold, $P=\sum_t{GT_t}$ is the total number of annotated objects among all frames, and $FP_r=\sum_t {FP_{r, t}}$ is the total number of false positive samples only considering the top confident samples that achieve the recall threshold $r$.
The hyperparameters $n=40$ and $\alpha=0.2$ (AMOTA@0.2), or $\alpha=1$ (AMOTA@1) are set by the benchmark organizers.
The overall AMOTA is the average AMOTA among all 7 categories.

\vspace{-5mm}

\subsection{Implementation details}
\lblsec{implementation}

Our implementation is based on CenterNet~\cite{zhou2019objects}.
We use DLA~\cite{yu2018deep} as the network backbone, optimized with Adam~\cite{kingma2014adam} with learning rate $1.25e-4$ and batchsize 32.
Data augmentations include random horizontal flipping, random resized cropping, and color jittering.
For all experiments, we train the networks for $70$ epochs. The learning rate is dropped by a factor of 10 at the 60th epoch.
We test the runtime on a machine with an Intel Core i7-8086K CPU and a Titan Xp GPU. The runtimes depend on the number of objects for rendering and the input resolution in each dataset.

The MOT dataset~\cite{MOT16} annotates each pedestrian as an amodal bounding box. That is, the bounding box always covers the whole body even when part of the object is out of the frame.
In contrast, CenterNet~\cite{zhou2019objects} requires the center of each inferred bounding box to be within the frame. 
To handle this, we separately predict the visible and amodal bounding boxes~\cite{tian2019fcos}.
Further details on this can be found in the supplement.
We follow prior works~\cite{yu2016poi,zhang2018integrated,sadeghian2017tracking,son2017multi,tang2017multiple} to pretrain on external data. 
We train our network on the CrowdHuman~\cite{shao2018crowdhuman} dataset, using the static image training described in Section~\ref{sec:trainimage}.
Details on the CrowdHuman dataset and ablations of pretraining are in the supplement.

The default input resolution for MOT images is ${1920 \times 1080}$.
We resize and pad the images to $960 \times 544$. 
We use random false positive ratio $\lambda_{fp}=0.1$ and random false negative ratio $\lambda_{fn}=0.4$.
We only output tracklets that have a confidence of $\theta=0.4$ or higher, and set the heatmap rendering threshold to $\tau=0.5$.
A controlled study of these hyperparameters is in the supplement.

For KITTI~\cite{Geiger2012CVPR}, we keep the original input resolution $1280 \times 384$ in training and testing. The hyperparameters are set at $\lambda_{fp}=0.1$ and $\lambda_{fn}=0.2$, with output threshold $\theta=0.4$ and rendering threshold $\tau=0.4$.
We fine-tune our KITTI model from a nuScenes tracking model.

For nuScenes~\cite{nuscenes2019}, we use input resolution ${800 \times 448}$. We set ${\lambda_{fp}=0.1}$ and ${\lambda_{fn}=0.4}$, and use output threshold ${\theta=0.1}$ and rendering threshold ${\tau=0.1}$.
We first train our nuScenes model for 140 epochs for just 3D detection~\cite{zhou2019objects} and then fine-tune for 70 epochs for 3D tracking.
Note that nuScenes evaluation is done per 360 panorama, not per image.
We naively fuse all outputs from the 6 cameras together, without handling duplicate detections at the intersection of views~\cite{simonelli2019disentangling}.

\paragraph{Track rebirth.}
\lblsec{rebirth}
Following common practice~\cite{zhang2018integrated,bergmann2019tracking},
we keep unmatched tracks ``inactive'' until they remain undetected for $K$ consecutive frames. 
Inactive tracks can be matched to detections and regain their ID, but not appear in the prior heatmap or output. 
The tracker stays online.
Rebirth only matters for the MOT test set, where we use $K=32$.
For all other experiments, we found rebirth not to be required ($K=0$).

\subsection{Public detection}
\lblsec{public}

The MOT17 challenge only supports public detection.
That is, participants are asked to use the provided detections.
Public detection is meant to test a tracker's ability to associate objects, irrespective of its ability to detect objects.
Our method operates in the private detection mode by default.
For the MOT challenge we created a public-detection version of CenterTrack that uses the externally provided (public) detections and is thus fairly compared to other participants in the challenge. This shows that the advantages of CenterTrack are not due to the accuracy of the detections but are due to the tracking framework itself.

Note that refining and rescoring the given bounding boxes is allowed and is commonly used by participants in the challenge~\cite{bergmann2019tracking,long2018tracking,keuper2018motion}.
Following Tracktor~\cite{bergmann2019tracking}, we keep the bounding boxes that are close to an existing bounding box in the previous frame.
We only initialize a new trajectory if it is near a public detection.
All bounding boxes in our results are either near a public detection in the current frame or near a tracked box in the previous frame.
The algorithm's diagram of this public-detection configuration can be found in the supplement.
We use this public-detection configuration of CenterTrack for MOT17 test set evaluation and use the private-detection setting in our ablation studies.

\begin{table}[t]
\center 
\footnotesize
\begin{tabular}{@{}l@{\ \ }c@{\ \ } c@{\ \ } c@{\ \ } c@{\ \ } c@{\ \ } c@{\ \ } c@{\ \ } c@{\ \ } c@{\ \ } c@{}}
\toprule
 & Time(ms) & MOTA $\uparrow$ & IDF1 $\uparrow$ & MT $\uparrow$ & ML $\downarrow$ & FP $\downarrow$ & FN $\downarrow$ & IDSW $\downarrow$\\
\midrule
Tracktor17~\cite{bergmann2019tracking} & 666+D & 53.5 & 52.3 & 19.5 & 36.6 & 12201 & 248047 & 2072 \\
LSST17~\cite{feng2019multi} & 666+D & 54.7 & \textbf{62.3}   & 20.4 & 40.1 & 26091 & 228434 & \textbf{1243} \\
Tracktor v2~\cite{bergmann2019tracking} & 666+D & 56.5 & 55.1   & 21.1 & 35.3 & \textbf{8866} & 235449 & 3763 \\
{GMOT} & {167+D} & {55.4} & {57.9}  & {22.7} & {34.7} & {20608} & {229511} & {1403} \\	
Ours (Public) & \textbf{57+D} & \textbf{61.5} & 59.6 & \textbf{26.4} & \textbf{31.9} & 14076 & \textbf{200672} & 2583\\
\midrule
Ours (Private) & 57 & 67.8 & 64.7 & 34.6 & 24.6 & 18498 & 160332 & 3039 \\
\bottomrule
\end{tabular}
\caption{Evaluation on the MOT17 test sets (top: public detection; bottom: private detection). We compare to published entries on the leaderboard. The runtime is calculated from the {HZ} column on the leaderboard. +D means detection time, which is usually $>100$ms~\cite{ren2015faster}.}
\lbltab{MOT}
\vspace{-7mm}
\end{table}

\subsection{Main results}
\lblsec{results}

All three datasets~-- MOT17~\cite{MOT16}, KITTI~\cite{Geiger2012CVPR}, and nuScenes~\cite{nuscenes2019}~-- host test servers with hidden annotations and leaderboards.
We compare to all published results on these leaderboards.
The numbers were accessed on Mar. 5th, 2020.
We retrain CenterTrack on the full training set with the same hyperparameters in the ablation experiments.

\reftab{MOT} lists the results on the MOT17 challenge.
We use our public configuration in \refsec{public} and do not pretrain on CrowdHuman~\cite{shao2018crowdhuman}.
CenterTrack significantly outperforms the prior state of the art even when restricted to the public-detection configuration. For example CenterTrack improves MOTA by 5 points (an 8.6\% relative improvement) over Tracktor v2~\cite{bergmann2019tracking}.

The public detection setting ensures that all methods build on the same underlying detector.
Our gains come from two sources.
Firstly, the heatmap input makes our tracker better preserve tracklets from the previous frame, which results in a much lower rate of false negatives.
And second, our simple learned offset is effective. (See \refsec{motion} for more analysis.)
For reference, we also included a private detection version, where CenterTrack simultaneously detects and tracks objects (\reftab{MOT}, bottom).
It further improves the MOTA to $67.3\%$, and runs at 17 FPS end-to-end (including detection).

For IDF1 and id-switch, our local model is not as strong as offline methods such as LSST17~\cite{feng2019multi}, but is better than other online methods~\cite{bergmann2019tracking}. We believe that there is an exciting avenue for future work in combining local trackers (such as our work) with stronger offline long-range models (such as SORT~\cite{Bewley2016_sort}, LMP~\cite{tang2017multiple}, and other ReID-based trackers~\cite{yu2016poi,xu2019spatial}).

\begin{table}[t]
\center 
\footnotesize
\begin{tabular}{@{}l@{\ \ }c@{\ \ }c@{\ \ }c@{\ \ }c@{\ \ }c@{\ \ }c@{\ \ }c@{\ \ }@{}}
\toprule
 & Time(ms) & MOTA $\uparrow$ & MOTP $\uparrow$ & MT $\uparrow$ & ML $\downarrow$ & IDSW $\downarrow$ & FRAG $\downarrow$ \\
\midrule
AB3D~\cite{Weng2019_3dmot} & 4+D & 83.84 & 85.24 & 66.92 & 11.38 & \textbf{9} & \textbf{224} \\
BeyondPixel~\cite{sharma2018beyond} & 300+D & 84.24 & 85.73 & 73.23 & 2.77 & 468 & 944\\
3DT~\cite{Hu3DT19} & 30+D & 84.52 & 85.64 & 73.38 & 2.77 & 377 & 847 \\
mmMOT~\cite{Zhang_2019_ICCV} & 10+D & 84.77 & 85.21 & 73.23 & 2.77 & 284 & 753 \\
MOTSFusion~\cite{luiten2019track} & 440+D & 84.83 & 85.21 & 3.08 & 2.77 & 275 & 759 \\
MASS~\cite{karunasekera2019multiple} & 10+D & 85.04 & \textbf{85.53} & 74.31 & 2.77 & 301 & 744 \\
Ours & 82 & \textbf{89.44} & 85.05 & \textbf{82.31} & \textbf{2.31} & 116 & 334 \\
\bottomrule
\end{tabular}
\caption{Evaluation on the KITTI test set. We compare to all published entries on the leaderboard. Runtimes are from the leaderboard. 
+D means detection time.}
\lbltab{sota:kitti}
\vspace{-1.25em}
\end{table}

\begin{table}[t]
\center 
\footnotesize
\begin{tabular}{@{}l@{\ \ }c@{\ \ }c@{\ \ }c@{\ \ }c@{\ \ }c@{}}
\toprule
 & Time(ms) & AMOTA@0.2 $\uparrow$ & AMOTA@1 $\uparrow$ & AMOTP $\downarrow$\\
Mapillary~\cite{simonelli2019disentangling}+AB3D~\cite{Weng2019_3dmot}  & - & 6.9 & 1.8 & 1.8 \\
Ours &45 & \textbf{27.8} &\textbf{4.6} & \textbf{1.5}\\
\bottomrule
\end{tabular}
\caption{Evaluation on the nuScenes test set. We compare to the official monocular 3D tracking baseline, which applies a state-of-the-art 3D tracker~\cite{Weng2019_3dmot}. We list the average AMOTA@0.2, AMOTA@1, and AMOTP over all 7 categories.}
\lbltab{sota:nuscenes}
\vspace{-7mm}
\end{table}

On KITTI~\cite{Geiger2012CVPR}, we submitted our best-performing model with flip testing~\cite{zhou2019objects}. 
The model runs at $82$ms and yields $89.44\%$ MOTA, outperforming all published work (\reftab{sota:kitti}).
Note that our model without flip testing runs at $45$ms with $88.7\%$ MOTA on the validation set (vs.\ $89.63\%$ with flip testing on the validation set). We avoid submitting to the test server multiple times following their test policy.
The results again indicate that CenterTrack performs competitively with more complex methods.

On nuScenes~\cite{nuscenes2019}, our monocular tracking method achieves an AMOTA@0.2 of $28.3\%$ and an AMOTA@1 of $4.6\%$, outperforming the monocular baseline~\cite{simonelli2019disentangling,Weng2019_3dmot} by a large margin.
There are two main reasons.
Firstly, we use a stronger and faster 3D detector~\cite{zhou2019objects} (see the 3D detector comparison in the supplementary).
More importantly, as shown in \reftab{motion}, the Kalman-filter-based 3D tracking baseline relies on hand-crafted motion rules~\cite{Weng2019_3dmot}, which are less effective in low-framerate regimes.
Our method learns object motion from data and is much more stable at low framerates.

\subsection{Ablation studies}
\lblsec{ablation}

\begin{table}[t]
{\center 
\small
\begin{tabular}{@{}l@{\ }c@{\ \ }c@{\ \ }c@{\ }c@{\ \ }c@{\ \ }c@{\ }c@{\ }c@{\ }c@{}c@{}}
\toprule
 & \multicolumn{4}{c}{MOT17} & \multicolumn{4}{c}{KITTI} & \multicolumn{2}{c}{nuScenes}\\
 & {\scriptsize MOTA$\uparrow$} & {\scriptsize FP$\downarrow$} & {\scriptsize FN$\downarrow$} & {\scriptsize IDSW$\downarrow$} & {\scriptsize MOTA$\uparrow$}  & {\scriptsize FP$\downarrow$} & {\scriptsize FN$\downarrow$} & {\scriptsize IDSW$\downarrow$} & {\scriptsize AMOTA@0.2$\uparrow$} & {\scriptsize AMOTA@1$\uparrow$} \\
\cmidrule(r){1-1}
\cmidrule(r){2-5}
\cmidrule(r){6-9}
\cmidrule(){10-11}
detection only & 63.6 & 3.5\% & 30.3\% & 2.5 \%
& 84.3  & 4.3\% & 9.8\% & 1.5\% & 18.1 & 3.4 \\
w/o offset & 65.8 & \textbf{4.5\%} & \textbf{28.4\%} & 1.3\% 
& 87.1 & \textbf{5.4\%} & \textbf{5.8\%} & 1.6\% & 17.8 & 3.6 \\
w/o heatmap & 63.9 & 3.5\% & 30.3\% & 2.3\%
& 85.4 & 4.3\% & 9.8\% & 0.4\% & 26.5 & 5.9\\
Ours & \textbf{66.1} & \textbf{4.5\%} & \textbf{28.4\%} & \textbf{1.0\%} 
& \textbf{88.7} & \textbf{5.4\%} & \textbf{5.8\%} & \textbf{0.1\%} & \textbf{28.3} & \textbf{6.8} \\
\bottomrule
\end{tabular}
\small
\caption{Ablation study on MOT17, KITTI, and nuScenes. All results are on validation sets (\refsec{datasets}). For each dataset, we report the corresponding official metrics. $\uparrow$~indicates that higher is better, $\downarrow$ indicates that lower is better.}
\lbltab{ablation}
}
\vspace{-7mm}
\end{table}

We first ablate our two main technical contributions: tracking-conditioned detection (\refsec{cond_track}) and offset prediction (\refsec{offset}) on all three datasets.
Specifically, we compare our full framework with three baselines.

\paragraph{Detection only} runs a CenterNet detector at each individual frame and associates their identity only based on 2D center distance. 
This model does not use video data, but still uses two input images.

\paragraph{Without offset} uses just tracking-conditioned prediction with a predicted offset of zero. Every object is again associated to its closest object in the previous frame.

\paragraph{Without heatmap} predicts the center offset between frames and uses the updated center distance as the association metric, but the prior heatmap is not provided. The offset-based greedy association is used.

\reftab{ablation} shows the results.
On all datasets, our full CenterTrack model performs significantly better than the baselines.
Tracking-conditioned detection yields $\sim2\%$ MOTA improvement on MOT and $\sim3\%$ MOTA improvement on KITTI, with or without offset prediction. 
It produces more false positives but fewer false negatives. 
This is because with the heatmap prior, the network tends to predict more objects around the previous peaks, which are sometimes misleading.
The merits of the heatmap outweigh the limitations and improve MOTA overall.
Using the prior heatmap also significantly reduces IDSW on both datasets, indicating that the heatmap stabilizes detection.

Tracking offset prediction gives a huge boost on nuScenes and reduces IDSW consistently in MOT and KITTI.
The effectiveness of the tracking offset appears to be related to the video framerate.
When the framerate is high, motion between frames is small, and a zero offset is often a reasonable starting point for association.
When framerate is low, as in the nuScenes dataset, motion between frames is large and static object association is considerably less effective. Our offset prediction scheme helps deal with such large inter-frame motion.
Next, we ablate other components on MOT17.

\paragraph{Training with noisy heatmap.} The 2nd row in \reftab{additional:mot} shows the importance of injecting noise into heatmaps during training (\refsec{train_video}). Without noise injection, the model fails to generalize and yields dramatically lower accuracy. In particular, this model has a large false negative rate. One reason is that in the first frame, the input heatmap is empty. This model had a hard time discovering new objects that were not indicated in the prior heatmap.

\paragraph{Training on static images.}
We train a version of our model on static images only, as described in \refsec{train_image}.
The results are shown in \reftab{additional:mot} (3rd row, `Static image'). 
As reported in this table, training on static images gives the same performance as training on videos on the MOT dataset.
Separately, we observed that training on static images is less effective on nuScenes, where framerate is low.

\begin{table}[t]
\center 
\footnotesize
\begin{tabular}{@{}l@{}c@{\ } c@{\ } c@{\ } c@{\ } c@{\ } c@{\ } c@{\ } c@{}}
\toprule
 & MOTA $\uparrow$ & IDF1 $\uparrow$ & MT $\uparrow$ & ML $\downarrow$ & FP $\downarrow$ & FN $\downarrow$ & IDSW $\downarrow$ \\
\midrule
Ours & 66.1 & 64.2 & 41.3 & 21.2 & 4.5\% & 28.4\% & 1.0\% \\
w.o. noisy hm & 34.4 & 46.2 & 26.3 & 42.2 & 7.3\% & 57.4\% & 0.9\% \\
Static image & 66.1 & 65.4 & 41.6 & 19.2 & 5.4\% & 27.5\% & 1.0\% \\
w. Hungarian & 66.1 & 61.0 & 40.7 & 20.9 & 4.5\% & 28.3\% & 1.0\% \\
w. rebirth & 66.2 & 69.4 & 39.5 & 22.1 & 3.9\% & 29.5\% & 0.4\% \\
\bottomrule
\end{tabular}
\caption{Additional experiments on the MOT17 validation set. From top to bottom: our model, our model trained without simulating heatmap noise, our model trained on static images only, our model with Hungarian matching, and our model with track rebirth.}
\lbltab{additional:mot}
\vspace{-7mm}
\end{table}

\paragraph{Matching algorithm.} We use a simple greedy matching algorithm based on the detection score, while most other trackers use the Hungarian algorithm. We show the performance of CenterTrack with Hungarian matching in the 4th row of \reftab{additional:mot}. It does not improve performance. We choose greedy matching for simplicity.

\paragraph{Track rebirth.} 
We show CenterTrack with track rebirth ($K$=$32$) in the last row of \reftab{additional:mot}. While the MOTA performance keeps similar, it significantly increases IDF1 and reduces ID switch. We use this setting for our MOT test set submission.
For other datasets and evaluation metrics no rebirth was required ($K=0$).

\begin{table}[t]
{\center 
\small
\begin{tabular}{@{}l@{\ }c@{\ \ }c@{\ \ }c@{\ }c@{\ \ }c@{\ \ }c@{\ }c@{\ }c@{\ }c@{}c@{}}
\toprule
 & \multicolumn{4}{c}{MOT17} & \multicolumn{4}{c}{KITTI} & \multicolumn{2}{c}{nuScenes}\\
 & {\scriptsize MOTA$\uparrow$} & {\scriptsize FP$\downarrow$} & {\scriptsize FN$\downarrow$} & {\scriptsize IDSW$\downarrow$} & {\scriptsize MOTA$\uparrow$}  & {\scriptsize FP$\downarrow$} & {\scriptsize FN$\downarrow$} & {\scriptsize IDSW$\downarrow$} & {\scriptsize AMOTA@0.2$\uparrow$} & {\scriptsize AMOTA@1$\uparrow$} \\
\cmidrule(r){1-1}
\cmidrule(r){2-5}
\cmidrule(r){6-9}
\cmidrule(){10-11}
no motion & 65.8 & \textbf{4.5\%} & \textbf{28.4\%} & 1.3\% 
& 87.1 & \textbf{5.4\%} & \textbf{5.8\%} & 1.6\% & 17.8 & 3.6 \\
Kalman filter & \textbf{66.1} & \textbf{4.5\%} & \textbf{28.4\%} & \textbf{1.0\%} 
& 87.9 & \textbf{5.4\%} & \textbf{5.8\%} & 0.9\% & 18.3 & 3.8 \\
optical flow &  \textbf{66.1} & \textbf{4.5\%} & \textbf{28.4\%} & \textbf{1.0\%} 
& 88.4 & \textbf{5.4\%} & \textbf{5.8\%} & 0.4\% & 26.6 & 6.2\\
ours & \textbf{66.1} & \textbf{4.5\%} & \textbf{28.4\%} & \textbf{1.0\%} 
& \textbf{88.7} & \textbf{5.4\%} & \textbf{5.8\%} & \textbf{0.1\%} & \textbf{28.3} & \textbf{6.8} \\
\bottomrule
\end{tabular}
\small
\caption{Comparing different motion models on MOT17, KITTI, and nuScenes. All results are on validation sets (\refsec{datasets}). All experiments on the same dataset are from the same model.}
\lbltab{motion}
}
\vspace{-7mm}
\end{table}

\vspace{-5mm}

\subsection{Comparison to alternative motion models}

\lblsec{motion}

Our offset prediction is able to estimate object motion, but also performs a simple association, as current objects are linked to prior detections, which CenterTrack receives as one of its inputs.
To verify the effectiveness of our learned association, we replace our offset prediction with three alternative motion models:

\paragraph{No motion.} We set the offset to zeros. It is copied from \reftab{ablation} for reference only.

\paragraph{Kalman filter.} The Kalman filter predicts each object's future state through an explicit motion model estimated from its history. It is the most widely used motion model in traditional real-time trackers~\cite{Bewley2016_sort,Wojke2017simple,Weng2019_3dmot}. We use the popular public implementation from SORT~\cite{Bewley2016_sort}.
%\footnote{\url{https://github.com/abewley/sort}}

\paragraph{Optical flow.} As an alternative motion model, we use FlowNet2~\cite{ilg2017flownet}. The model was trained to estimate dense pixel motion for all objects in a scene. We run the strongest officially released FlowNet2 model ($\sim 150$ms / image pair), and replace our learned offset with the predicted optical flow at each predicted object center.
%\footnote{\url{https://github.com/NVIDIA/flownet2-pytorch}}

The results are shown in \reftab{motion}.
All models use the exact same detector.
On the high-framerate MOT17 dataset, any motion model suffices, and even no motion model at all performs competitively.
On KITTI and nuScenes, where the intra-frame motions are non-trivial, the hand-crafted motion rule of the Kalman filter performs significantly worse, and even the performance of optical flow degrades.
This emphasizes that our offset model does more than just motion estimation.
CenterTrack is conditioned on prior detections and can learn to snap offset predictions to exactly those prior detections.
Our training procedure strongly encourages this through heavy data augmentation.

{
\begin{figure}[t]
\centering
     \begin{tabular}{c@{\ }c@{\ }c}
     \includegraphics[width=0.33\textwidth]{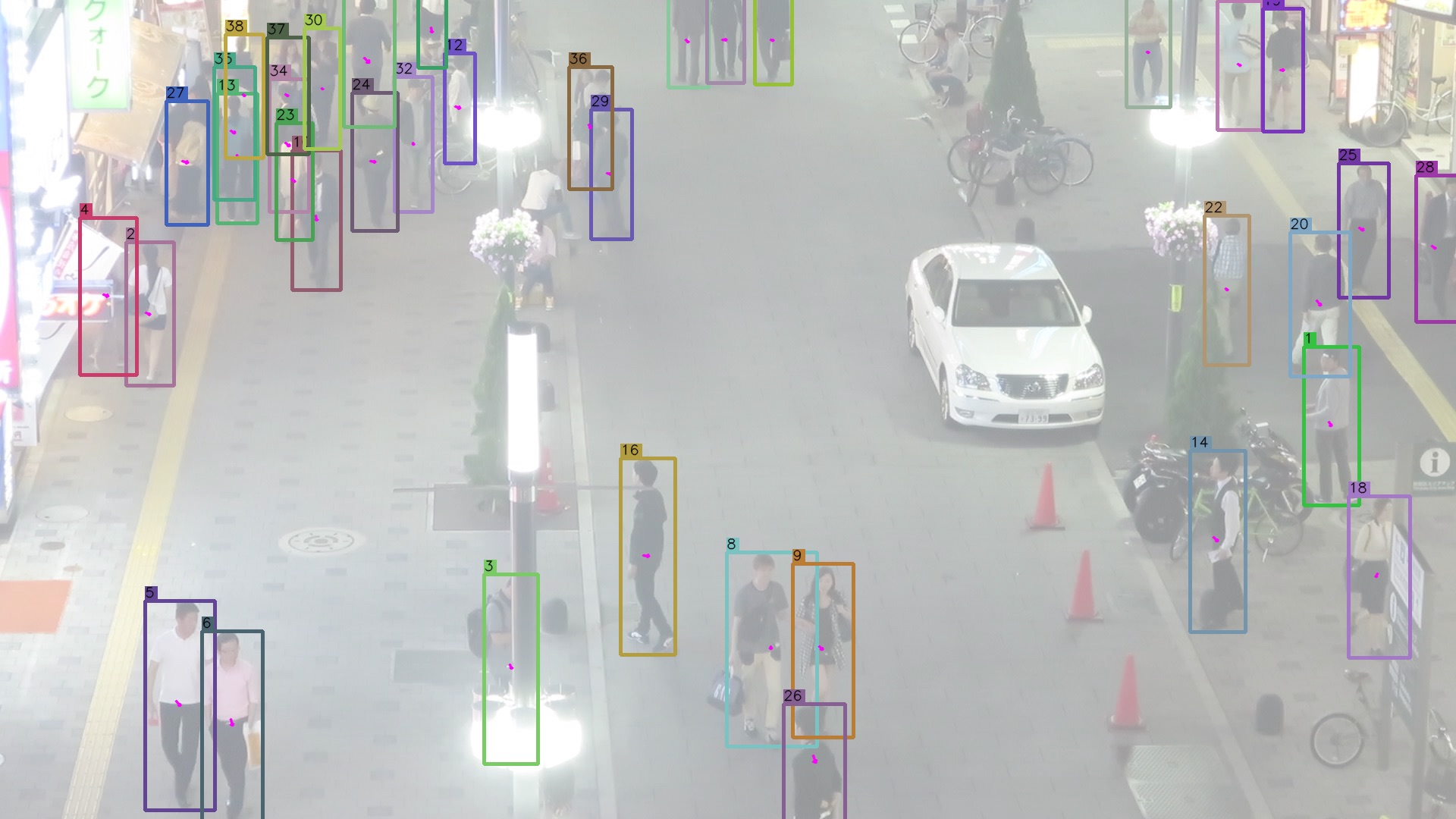}
      &\includegraphics[width=0.33\textwidth]{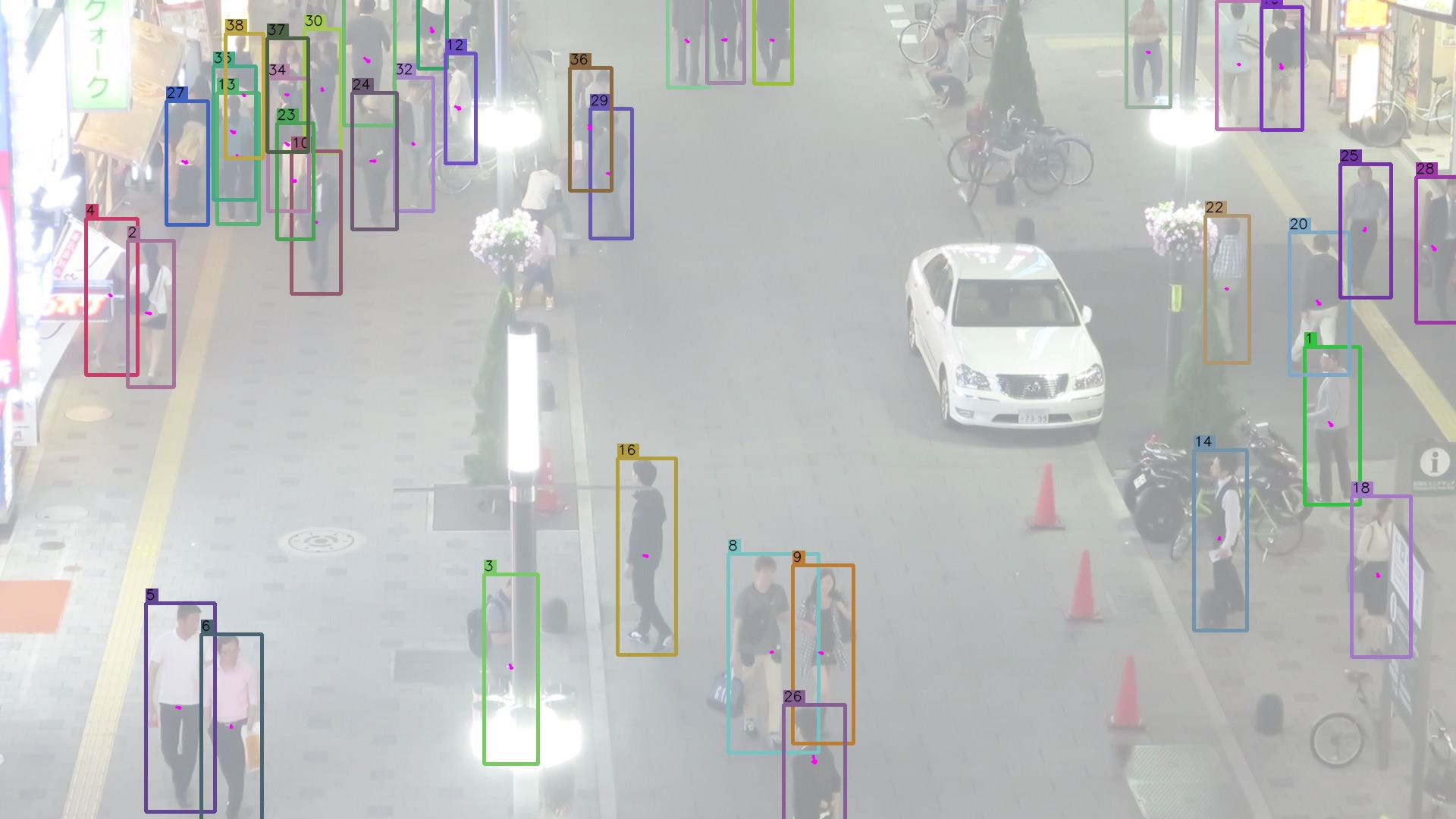}
      &\includegraphics[width=0.33\textwidth]{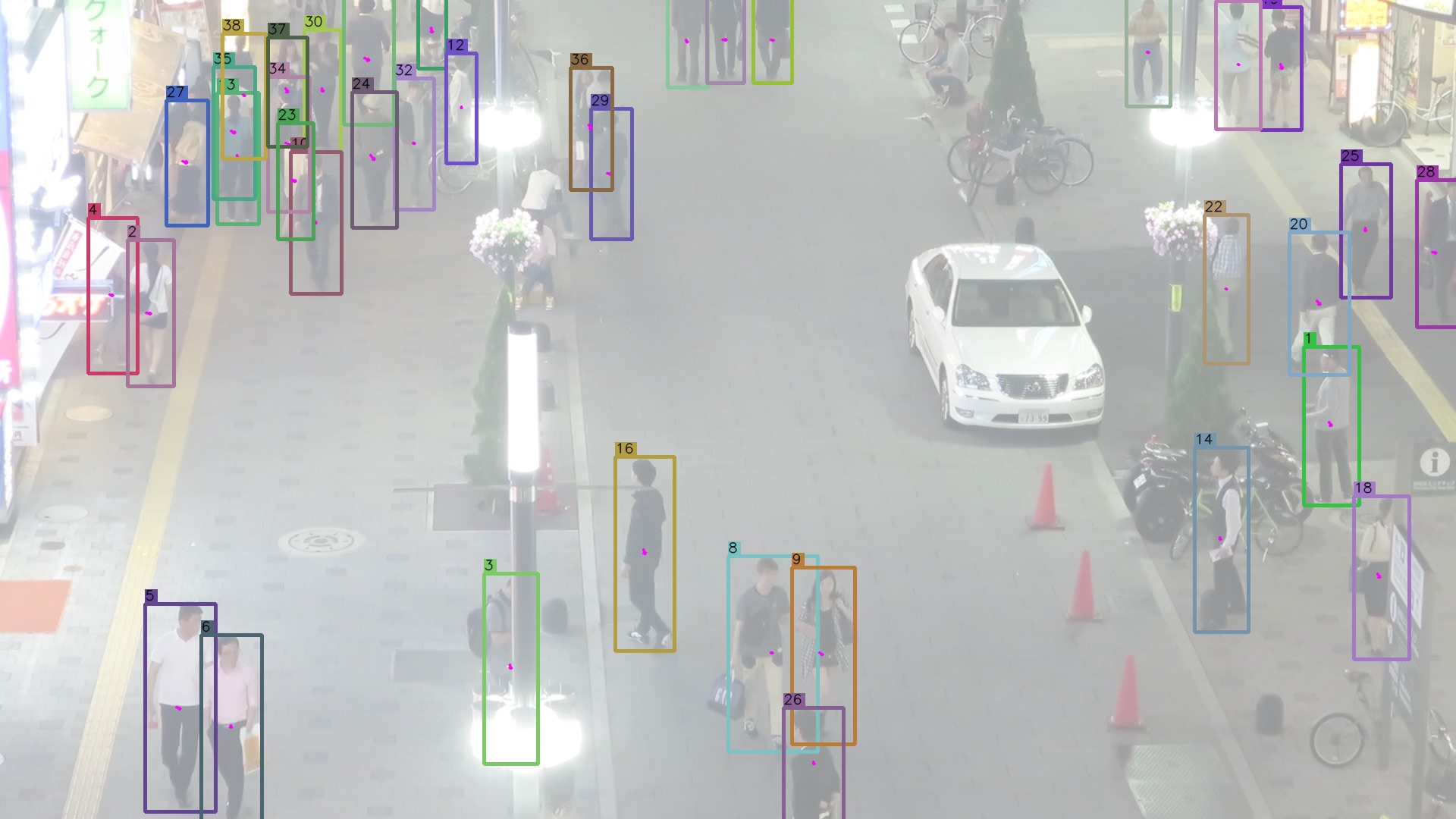} \\
     \includegraphics[width=0.33\textwidth]{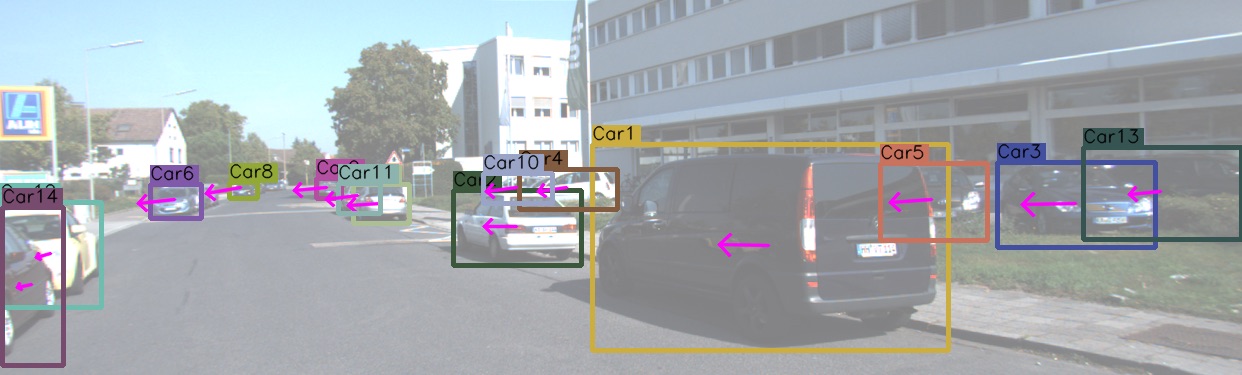}
      &\includegraphics[width=0.33\textwidth]{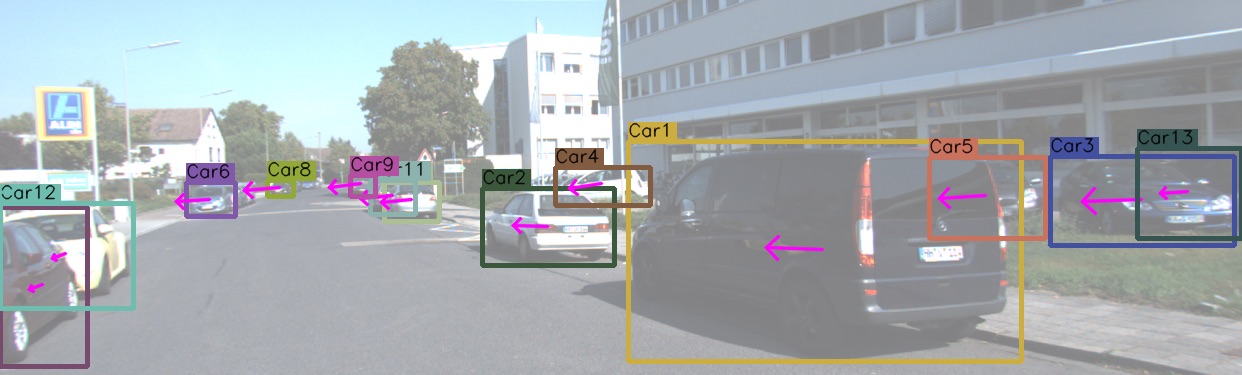}
      &\includegraphics[width=0.33\textwidth]{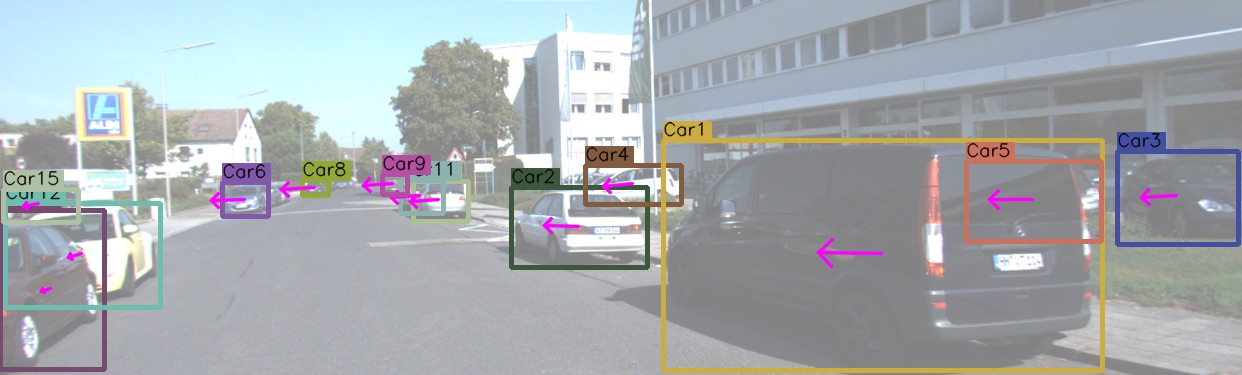} \\
     \includegraphics[width=0.33\textwidth]{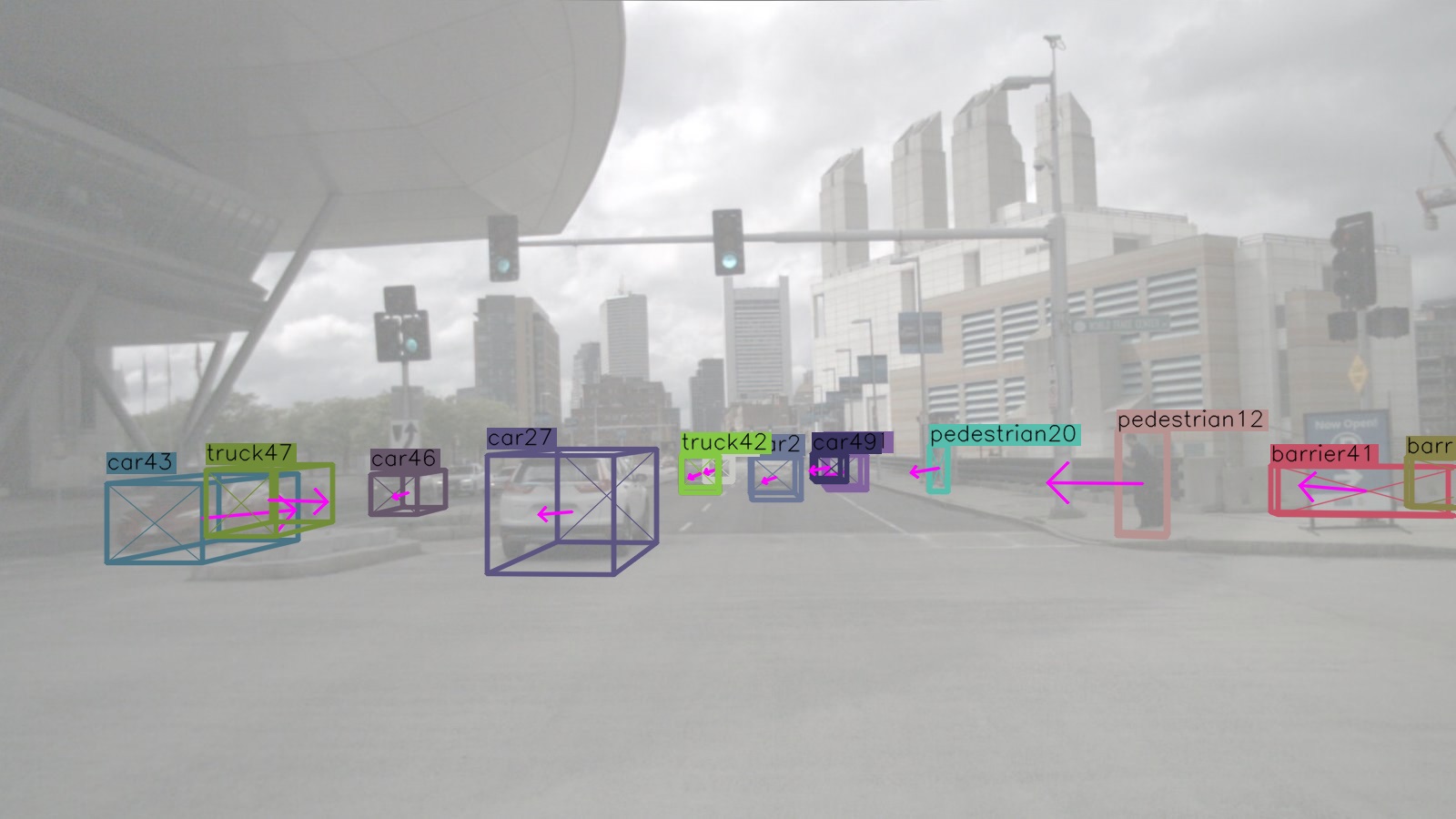}
      &\includegraphics[width=0.33\textwidth]{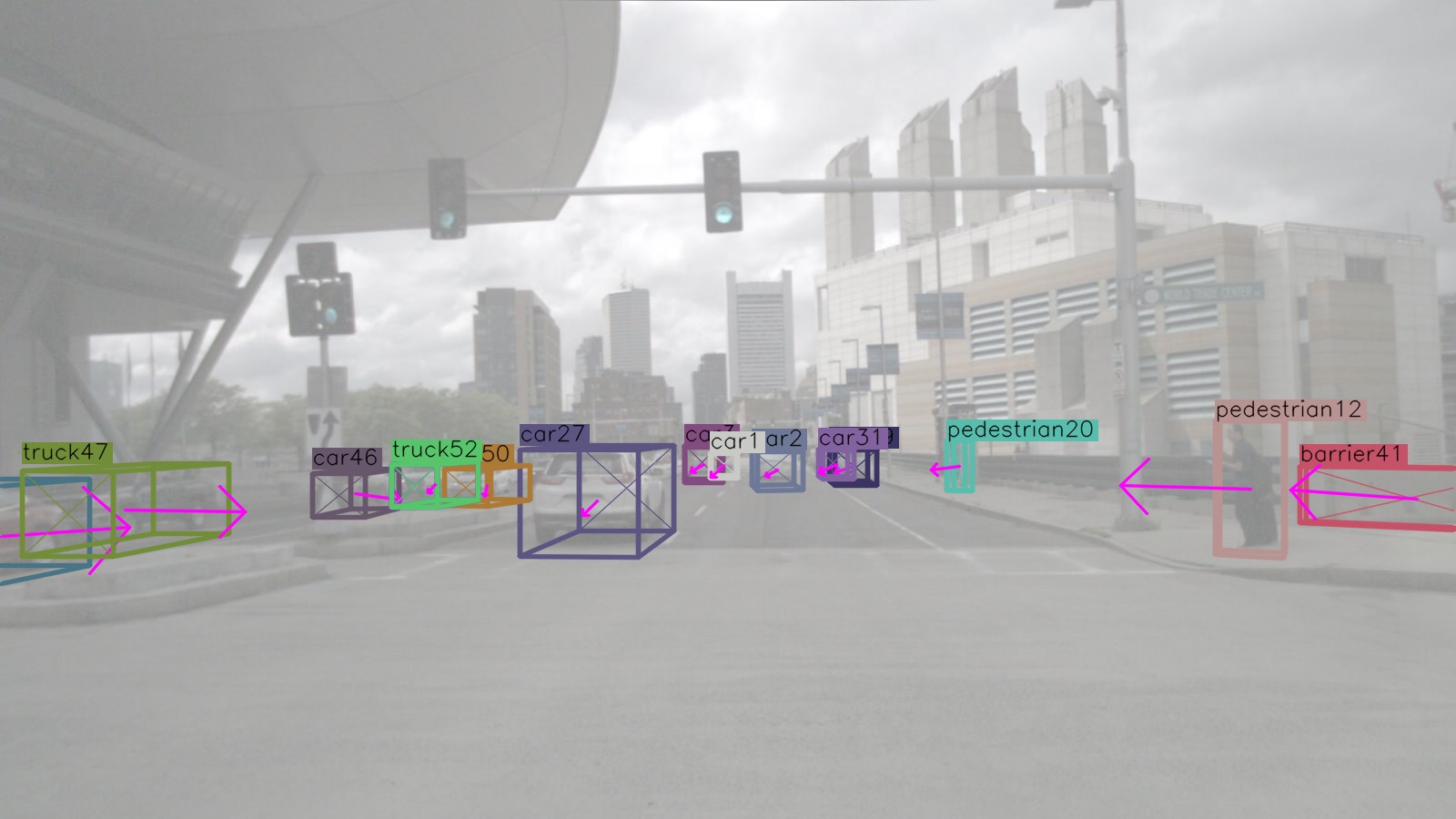}
      &\includegraphics[width=0.33\textwidth]{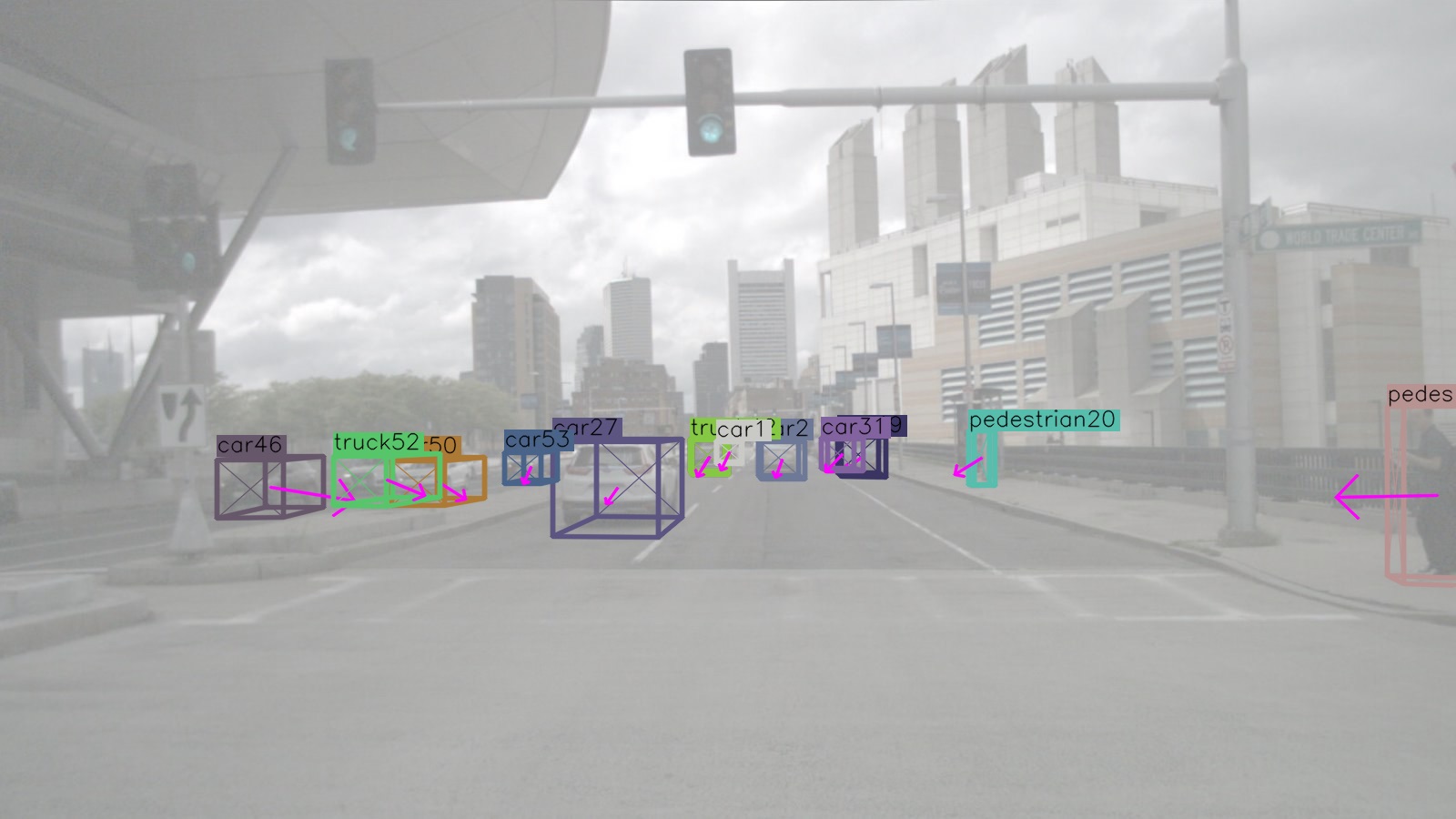}\\
     \includegraphics[width=0.33\textwidth]{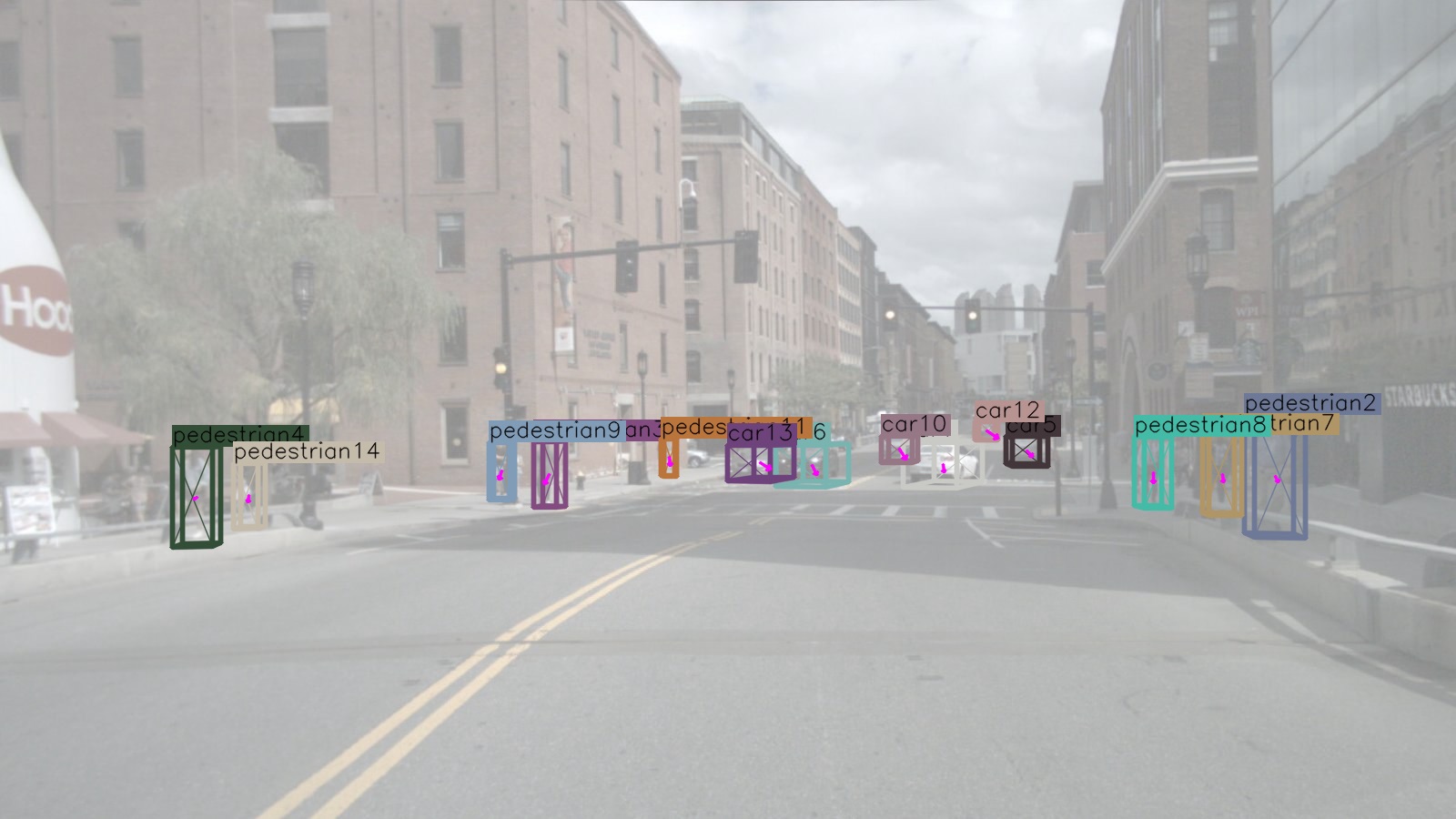}
      &\includegraphics[width=0.33\textwidth]{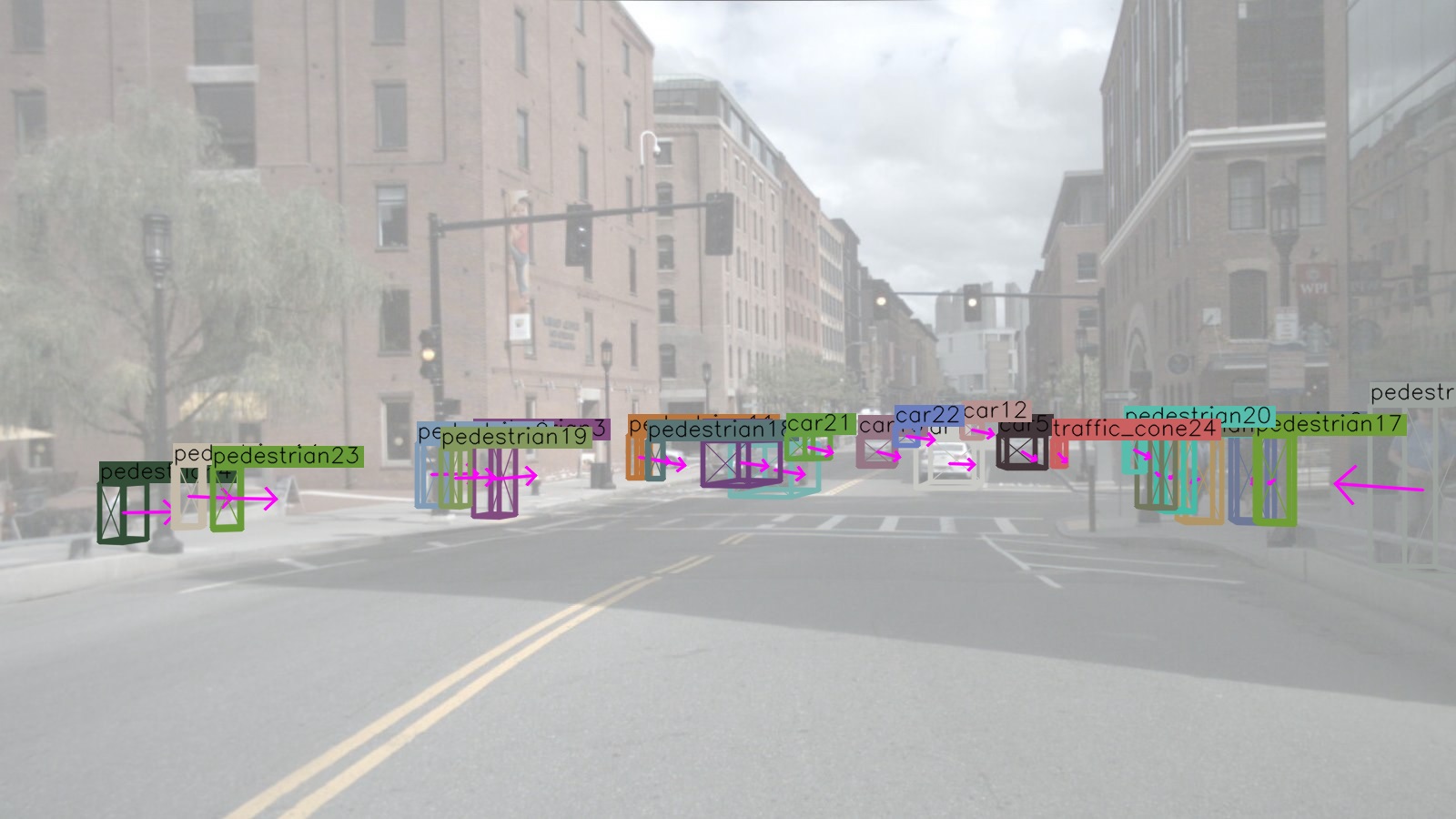}
      &\includegraphics[width=0.33\textwidth]{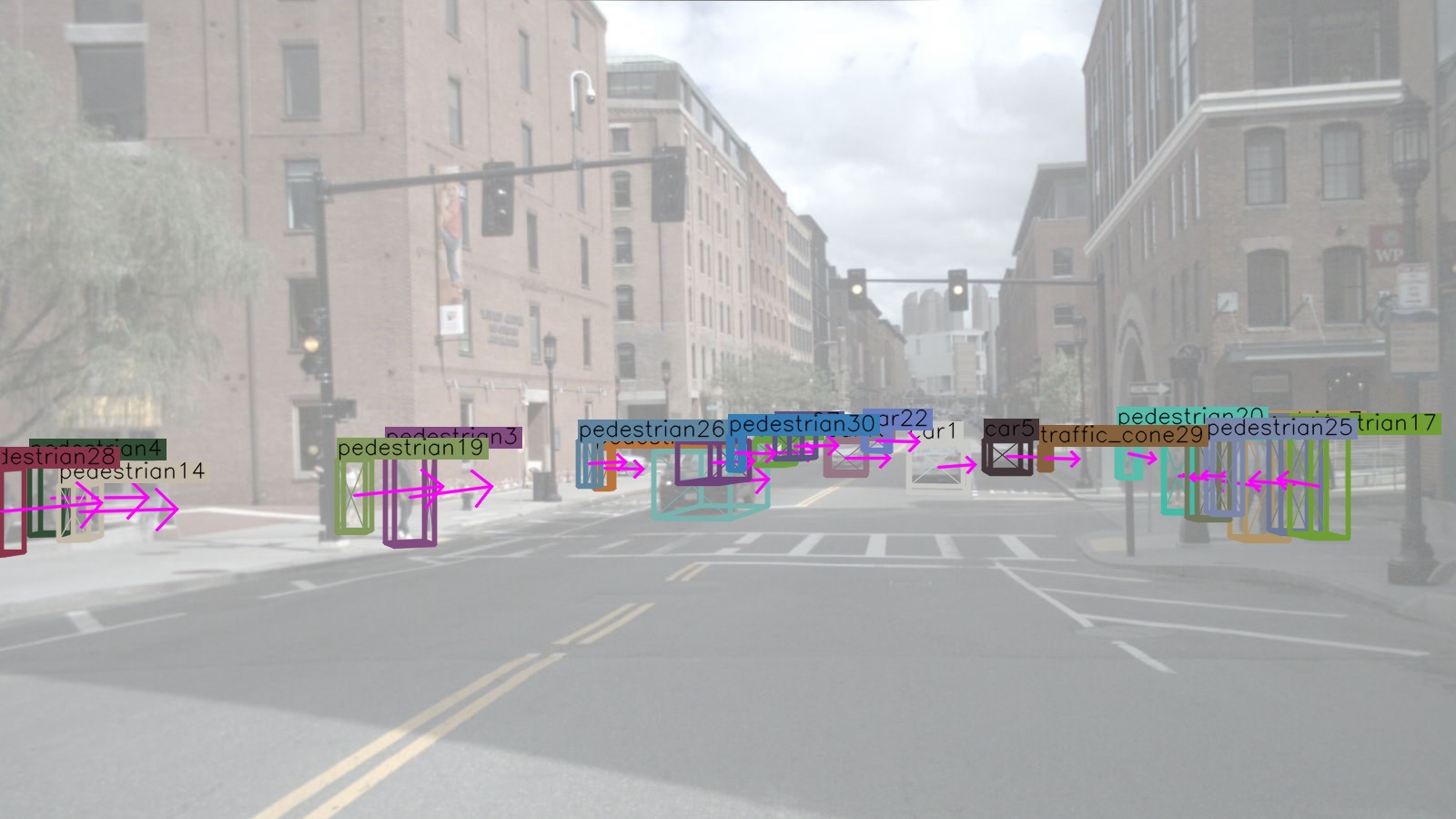} \\
      \end{tabular}
      \caption{Qualitative results on MOT (1st row), KITTI (2nd row), and nuScenes (3rd and 4th rows). Each row shows three consecutive frames. We show the predicted tracking offset in arrow. Tracks are coded by color. Best viewed on the screen.}
      \label{fig:demo}
\vspace{-7mm}
\end{figure}
}

\section{Conclusion}

We presented an end-to-end simultaneous object detection and tracking framework. Our method takes two frames and a prior heatmap as input, and produces detection and tracking offsets for the current frame. 
Our tracker is purely local and associates objects greedily through time.
It runs online (no knowledge of future frames) and in real time, and sets a new state of the art on the challenging MOT17, KITTI, and nuScenes 3D tracking benchmarks.

\paragraph{Acknowledgements.}
This work has been supported in part by the National Science Foundation under grant IIS-1845485.

\bibliographystyle{splncs04}
\bibliography{egbib}

\clearpage

\appendix

\section{Tracking algorithms}
\subsection{Private tracking}
We adopt a simple greedy id association algorithm based on the center distance, shown in Algorithm \ref{alg:association_pri}.
We use the same algorithm for both 2D tracking and 3D tracking.

\subsection{Public tracking}

For public tracking, we follow Tractor~\cite{bergmann2019tracking} to extend a private tracking algorithm to public detection. The id association is exactly the same as private detection (Line \ref{alg:st} to Line \ref{alg:ed}). The difference lies in how a track can be created. In public detection, we only initialize a track if it is near a provided bounding box (Line \ref{alg:pub:st} to Line \ref{alg:pub:ed}).

\section{Results on MOT16}
MOT16 shares the same training and testing sequences with MOT17, but officially supports private detection.
As is shown in \reftab{sota:MOT}, we rank 2nd among all published entries.
We remark that all other entries use a heavy detector trained on private data~\cite{yu2016poi} and many rely on slow matching schemes~\cite{tang2017multiple,yu2016poi}.
For example, LMP\_p~\cite{tang2017multiple} computes person-reidentification features for all pairs of bounding boxes using a Siamese network, requiring $O(n^2)$ forward passes through a deep network. In contrast, CenterTrack involves a single pass through a network and operates online at 17 FPS.

\begin{table}[b]
\center \
\begin{tabular}{@{}l@{}c@{\ } c@{\ } c@{\ } c@{\ } c@{\ } c@{\ } c@{\ } c@{}}
\toprule
 & Time(ms) & MOTA $\uparrow$ & IDF1 $\uparrow$ & FP $\downarrow$ & FN $\downarrow$ & IDSW $\downarrow$\\
\midrule
SORT~\cite{Bewley2016_sort} & 36+D & 60.4 & 56.1  & 11183 & 59867 & 1135 \\
DeepSORT~\cite{Wojke2017simple} & 59+D & 61.4 & 62.2  & 12852 & 56668 & 781\\
POI~\cite{yu2016poi} & 100+D & 66.1 & 65.1 & 5061 & 55914 & 805 \\
KNDT~\cite{yu2016poi} & 1428+D & 68.2 & 60.0 & 11479 & 45605 & 933 \\	
LMP\_p ~\cite{tang2017multiple} & 2000+D & \textbf{71.0} & \textbf{70.1} & \textbf{7880} & 44564 & \textbf{434} \\	
Ours (Private) & \textbf{57} & 69.6 & 60.7 & 10458 & \textbf{42805} & 2124 \\
\bottomrule
\end{tabular}
\normalsize
\caption{Evaluation on the MOT16 test sets (private detection). We compare to all published on the leaderboard. The runtime is calculated from the {HZ} column on the leaderboard. +D means detection time, which is usually $>100$ms~\cite{ren2015faster}.}
\lbltab{sota:MOT}
\end{table}

{
\hspace{-10mm}
\begin{minipage}{0.52\textwidth}
\small
\begin{algorithm}[H]
	\caption{\small{Private Detection}}
	\label{alg:association_pri}
	\setstretch{1.158}
	\SetAlgoLined
	\SetKwInOut{Input}{Input} \SetKwInOut{Output}{Output}
    \Input{$T^{(t - 1)} = \{(\vec{p}, \vec{s}, id)_j^{(t-1)}\}_{j=1}^{M}$:
    Tracked objects in the previous frame, with center $\vec p$, size $\vec s = (w, h)$. 
    $\hat{B}^{(t)} = \{(\hat{\vec p}, \hat{\vec d})_i^{(t)}\}_{i=1}^{N}$: Heatmap peaks with offset $\hat{\vec d}$ in the current frame, sorted in desending confidence.
    \ }
	\Output{$T^{(t)} = \{(\vec{p}, \vec{s}, id)_i^{(t)}\}_{i=1}^{N}$: Tracked objects in the current frame.}
	// \textbf{Initialization:} $T^{(t)}$ and $\mathcal{S}$ are initialized as empty lists. \label{alg:st} \\
	$T^{(t)} \leftarrow \emptyset$ \\
	$S \leftarrow \emptyset$ \ \ // Set of matched tracks \\
	$W \leftarrow Cost(B^{(t)}, T^{(t-1)})$// $W_{ij} = ||\hat{\vec{p}_i}^{(t)} - \hat{\vec{d}_i}^{(t)}, \vec{p}_j^{(t-1)}||_2$ \\
	\ \\
	\For{$i \leftarrow 1 \ to \ N$} {
		 $j \leftarrow \argmin_{j \notin {S}} W_{ij}$ \\
		 // calculate the distance threshold $\kappa$ \\ 
		 $\kappa \leftarrow \min(\sqrt{\hat{w}_i \hat{h}_i}, \sqrt{w_j h_j})$ \\
		 // if the cost is smaller the threshold. \\
		 \If {$w_{ij} < \kappa$} {
		    // Propagate matched id \\
		    $T^{(t)} \leftarrow T^{(t)} \cup (\hat{\vec{p}}_i^{(t)}, \hat{\vec{s}}_i^{(t)}, id_j^{(t-1)})$ \\
		    $ S \leftarrow S \cup \{j\}$ // Mark track j as matched \label{alg:ed} \\
		 } \Else {
		    \ \\ 
		    \ \\
		    \ \\
		    // Create a new track. \\
		    $T^{(t)} \leftarrow T^{(t)} \cup (\hat{\vec{p}}_i^{(t)}, \hat{\vec{s}}_i^{(t)}, NewId)$ \\
		    \ \\
		 }
	}
	\textbf{Return:} $T^{(t)}$
\end{algorithm}
\end{minipage}
\hfill
\hspace{2mm}
\begin{minipage}{0.52\textwidth}
\small
\begin{algorithm}[H]
	\caption{\small{Public Detection}}
	\label{alg:association_pub}
	\SetAlgoLined
	\SetKwInOut{Input}{Input} \SetKwInOut{Output}{Output} 
    \Input{$T^{(t - 1)} = \{(\vec{p}, \vec{s}, id)_j^{(t-1)}\}_{j=1}^{M}$: Tracked objects in the previous frame, with center $\vec p$, size $\vec s = (w, h)$.
    $\hat{B}^{(t)} = \{(\hat{\vec p}, \hat{\vec d})_i^{(t)}\}_{i=1}^{N}$: Heatmap peaks with offset $\hat{\vec d}$ in the current frame, sorted in desending confidence.
    $\hat{D}^{(t)} = \{(\vec{p}, \vec{s})_k^{(t)}\}_{k=1}^K$: Public detections. }
	\Output{$T^{(t)} = \{(\vec{p}, \vec{s}, id)_{i'}^{(t)}\}_{i'=1}^{N'}$: Tracked objects in the current frame.}
	// \textbf{Initialization:} $T^{(t)}$ and $\mathcal{S}$ are initialized as empty lists. \\
	$T^{(t)} \leftarrow \emptyset$ \\
	$S \leftarrow \emptyset$ \ \ // Set of matched tracks \\
	$W \leftarrow Cost(B^{(t)}, T^{(t-1)})$// $W_{ij} = ||\hat{\vec{p}_i}^{(t)} - \hat{\vec{d}_i}^{(t)}, \vec{p}_j^{(t-1)}||_2$ \\
	$W' \leftarrow Cost(B^{(t)}, D^{(t)})$// $W'_{ik} = ||\hat{\vec{p}_i}^{(t)}, \vec{p}_k^{(t)}||_2$ \\
	\For{$i \leftarrow 1 \ to \ N$} {
		 $j \leftarrow \argmin_{j \notin {S}} W_{ij}$ \\
		 // calculate the distance threshold $\kappa$ \\ 
		 $\kappa \leftarrow \min(\sqrt{\hat{w}_i \hat{h}_i}, \sqrt{w_j h_j})$ \\
		 // if the cost is smaller the threshold. \\
		 \If {$w_{ij} < \kappa$} {
		    // Propagate matched id \\
		    $T^{(t)} \leftarrow T^{(t)} \cup (\hat{\vec{p}}_i^{(t)}, \hat{\vec{s}}_i^{(t)}, id_j^{(t-1)})$ \\
		    $ S \leftarrow S \cup \{j\}$ // Mark track j as matched \\
		 } \Else {
		    $k \leftarrow \argmin_{k=1}^{K} \leftarrow W'_{ik}$ \label{alg:pub:st} \\
		    $\kappa' \leftarrow \min(\sqrt{\hat{w}_i \hat{h}_i}, \sqrt{w_k h_k})$ \\
		    \If {$W'_{ik} < \kappa'$} {
		        // Create a new track. \\
		        $T^{(t)} \leftarrow T^{(t)} \cup (\hat{\vec{p}}_i^{(t)}, \hat{\vec{s}}_i^{(t)}, NewId)$ \\
		    \label{alg:pub:ed} } 
		 }
	}
	
	\textbf{Return:} $T^{(t)}$
\end{algorithm}
\end{minipage}
}

\begin{table}[t]
    \begin{center}
    \begin{tabular} {l c c c c c c c c}
        \toprule
        & {Modality} & {mAP} $\uparrow$ & {mATE} $\downarrow$ & {mASE} $\downarrow$ & {mAOE} $\downarrow$ & {mAVE} $\downarrow$ & {mAAE} $\downarrow$ & {NDS} $\uparrow$ \\
        \midrule
        Megvii \cite{zhu2019class} & LiDAR & 52.8 & {0.300} & 0.247 & {0.379} & {0.245} & 0.140 & {63.3} \\
        PointPillars \cite{lang2019pointpillars} & LiDAR & 30.5 & {0.517} & 0.290 & {0.500} & {0.316} & 0.368 & {45.3} \\
        Mappilary \cite{simonelli2019disentangling} & Camera & 30.4 & 0.738 & 0.263 & 0.546 & 1. & {0.134} & 38.4 \\
        CenterNet~\cite{zhou2019objects} & Camera & {33.8} & 0.658 & {0.255} & 0.629 & 1. & 0.141 & 40.1 \\
        \bottomrule
    \end{tabular}
    \end{center}
    \caption{3D detection results on nuScenes test set. We show 3D bounding box mAP, mean translation error (mATE), mean size error (mASE), mean orientation error (mAOE), mean velocity error (mATE), mean attributes error (mAAE), and their weighted (with weight 5 on mAP and 1 on others) average NDS.}
    \vspace{-5mm}
    \lbltab{nuscenesdet}
\end{table}

\section{3D detection}

We follow CenterNet~\cite{zhou2019objects} to regress to object depth $\hat{D} \in R^{\frac{W}{R} \times \frac{H}{R}}$, 3d extent $\hat{\Gamma} \in R^{\frac{W}{R} \times \frac{H}{R} \times 3}$, orientation (encoded as an 8-dimension vector) $\hat{A} \in R^{\frac{W}{R} \times \frac{H}{R} \times 8}$. The training loss for these are identical to CenterNet~\cite{zhou2019objects}.
Since the 2D bounding box center does not align with the projected 3D bounding box center due to perspective projection, we in addition regress to an offset from the 2D center to the projected 3D bounding box center$\hat{F} \in R^{\frac{W}{R} \times \frac{H}{R} \times 2}$. We use L1Loss:
\begin{equation}
    L_{off3d} = \frac{1}{N}\sum_{k = 1}^N{|\hat{f}_k - f_k|},
\end{equation}

\noindent where $f_k \in \mathcal{R}^2$ is the ground truth offset of object $k$, and $\hat{f}_k = \hat{F}_{\vec{p}_k}$ is the value in $\hat{F}$ at location $\vec{p}_k$. 

We show the 3D detection performance of CenterNet~\cite{zhou2019objects} with the offset prediction in \reftab{nuscenesdet} for reference. The 3D detection performance is on-par with Mappilary~\cite{simonelli2019disentangling} and PointPillars~\cite{lang2019pointpillars}, but far below the LiDAR based state-of-the-art Megvii~\cite{zhu2019class}.

\section{Amodal bounding box regression}

CenterNet~\cite{zhou2019objects} requires the bounding box center to be within the image.
While in MOT~\cite{MOT16}, the center of the annotated bounding box (Amodal bounding box) can be outside of the image. To accommodate this case, We extend the $2$-channel bounding 
box size head in CenterNet to a $4$-channel head $\hat{A} \in \mathbb{R}^{\frac{W}{R} \times \frac{H}{R} \times 4}$ for the distance to the top-, left-, bottom-, right-bounding box border. 
Note that we still detect the in-frame bounding box center and regress to the in-frame bounding box size.
With this $4$-dimensional bounding box formulation, the output bounding box is not necessarily centered on the detected center. The training loss for the $4$-dimensional bounding box formulation is L1Loss:
\begin{equation}
    L_{amodal\_size} = \frac{1}{N}\sum_{i = 1}^N{|\hat{A}_{p_i} - a_i|}
\end{equation}
where $a_i \in \mathbb{R}^4$ is the ground truth border distance.

\begin{table}[b]
\center 
\begin{tabular}{@{}l@{}c@{\ } c@{\ } c@{\ } c@{\ } c@{\ } c@{\ } c@{\ } c@{}}
\toprule
 & MOTA $\uparrow$ & IDF1 $\uparrow$ & MT $\uparrow$ & ML $\downarrow$ & FP $\downarrow$ & FN $\downarrow$ & IDSW $\downarrow$ \\
\midrule
Ours & 66.1 & 64.2 & 41.3 & 21.2 & 4.5\% & 28.4\% & 1.0\% \\
only CrowdH. & 52.2 & 53.8 & 33.6 & 25.1 & 6.7\% & 39.7\% & 1.4\% \\
scratch & 60.7 & 62.8 & 33.0  & 22.4 & 4.0\% & 34.2\% & 1.0\% \\
scratch-Pub. & 57.4 & 59.6 & 31.1 & 27.1 & 2.1\% & 39.6\% & 1.0\% \\
\bottomrule
\end{tabular}
\caption{Additional experiments on the MOT17 validation set. From top to bottom: our full model, the model trained only on CrowdHuman dataset, our model trained from scratch, and the public detection mode of our model trained from scratch.}
\lbltab{mot}
\vspace{-5mm}
\end{table}

\begin{table}
\center 
\begin{tabular}{@{}l@{}c@{\ } c@{\ } c@{\ } c@{\ } c@{\ } c@{\ } c@{}}
\toprule
 & MOTA $\uparrow$ & MOTP $\uparrow$ & MT $\uparrow$ & ML $\downarrow$ & FP $\downarrow$ & FN $\downarrow$ & IDSW $\downarrow$ \\
\midrule
Ours & 88.7 & 86.7 & 90.3 & 2.1 & 5.4\% & 5.8\% & 0.1\% \\
Static image & 86.8 & 86.5 & 88.5 & 2.2 & 4.8\% & 7.9\% & 0.4\% \\
w.o. noisy hm & 80.1 & 85.3 & 76.2 & 7.6 & 3.8\% & 16.1\% & 0.1\% \\
Hungarian & 88.7 & 86.7 & 90.3 & 2.1 & 5.4\% & 5.8\% & 0.1\% \\
scratch & 84.5 & 83.2 & 83.4 & 2.8 & 5.7\% & 9.6\% & 0.3\% \\
\bottomrule
\end{tabular}
\caption{Additional experiments on the KITTI validation set. From top to bottom: our full model, the public-detection configuration of our model, our model trained on static images only, our model trained without simulating heatmap noise, our model with the Hungarian algorithm used for matching, and our model trained from scratch.}
\vspace{-5mm}
\lbltab{kitti}
\end{table}

\section{CrowdHuman dataset}

CrowdHuman~\cite{shao2018crowdhuman}contains 15k training images with common pose annotations.
The dataset is featured of high density and large occlusion. Both visible bounding box and the Amodal bounding box are annotated. We use the Amodal bounding box annotation in our experiments to align with MOT~\cite{MOT16}.

\section{Pretraining experiments}

For pretraining on CrowdHuman~\cite{shao2018crowdhuman}, we use input resolution $512 \times 512$, false positive ratio $\lambda_{fp} = 0.1$, false negative ratio $\lambda_{fn} = 0.4$, random scaling ratio $0.05$, and random translation ratio $0.05$. The training follows Section.4.4 of the main paper. As shown in \reftab{mot}, the model trained on CrowdHuman achieves a decent $52.2$ MOTA in MOT dataset, without seeing any MOT data.

Without CrowdHuman~\cite{shao2018crowdhuman} pretraining, our performance drops to $60.7 \%$ MOTA on the validation set. Pretraining help improve detection quality by decreasing the false negatives. Note that most entries on MOT challenges use external data for pretraining, and some of them use private data~\cite{yu2016poi}.
For reference, we also show our public detection results without pretraining in \reftab{mot}, last row. This model corresponds to the entry we submitted to MOT17 public detection challenge.

\section{Additional experiments on KITTI}

In \reftab{kitti}, we show results of the same additional experiments (Section. 5.5 of the main paper) on KITTI dataset~\cite{Geiger2012CVPR}. The conclusions are the same as on MOT~\cite{MOT16}.
Training on static images now performs slightly worse than training on video, mostly due to that KITTI has larger inter-frame motion than MOT. Training without random heatmap noise is much worse than the full model, with a high false-negative rate. And using the Hungarian algorithm works the same as using a greedy matching.
Our model without nuScenes~\cite{nuscenes2019} achieves $84.5\%$ MOTA on the validation set, this is on-par with other state-of-the-art trackers on KITTI~\cite{Hu3DT19,sharma2018beyond,Weng2019_3dmot} with a heavy detector~\cite{ren2017accurate}.

\begin{table}[t]
\center 
\begin{tabular}{@{}c@{\ \ \ \ }c@{\ }c@{\ } c@{\ } c@{\ } c@{\ } c@{\ } c@{\ } c@{\ } c@{}}
\toprule
$\theta$ & $\tau$ & MOTA $\uparrow$ & IDF1 $\uparrow$ & MT $\uparrow$ & ML $\downarrow$ & FP $\downarrow$ & FN $\downarrow$ & IDSW $\downarrow$ \\
\midrule
0.4 & 0.4 & 62.6 & 64.9 & 44.0 & 18.9 & 10.3\% & 26.4\% & 0.7\% \\
0.4 & 0.6 & 65.5 & 63.2 & 38.6 & 22.4 & 2.5\% & 30.5\% & 1.5\% \\
0.4 & 0.5 & 66.1 & 64.2 & 41.3 & 21.2 & 4.5\% & 28.4\% & 1.0\% \\
0.3 & 0.5 & 66.2 & 64.3 & 43.1 & 19.2 & 5.7\% & 26.9\% & 1.2\% \\
0.5 & 0.5 & 65.2 & 62.1 & 39.8 & 23.0 & 3.7\% & 30.2\% & 0.9\% \\
\bottomrule
\end{tabular}
\caption{Experiments with different output thresholds ($\theta$) and rendering thresholds ($\tau$) on the MOT~\cite{MOT16} validation set. We search $\theta$ and $\tau$ locally in a step of $0.1$.}
\lbltab{thresholds}
\vspace{-5mm}
\end{table}

\section{Output and rendering threshold}

As the tracking evaluation metric (MOTA) does not consider the confidence of predictions, picking an output threshold is essential in all tracking algorithms (see discussion in AB3D~\cite{Weng2019_3dmot}). In our case, we also need a threshold to render predictions to the prior heatmap. We search the optimal thresholds on MOT~\cite{MOT16} in \reftab{thresholds}. Basically, increasing both thresholds results in fewer outputs, thus increases the false negatives while decreases the false positives. We find a good balance at $\theta=0.4$ and $\tau=0.5$. 

\end{document}